\pdfoutput=1
\documentclass[10.5pt, journal]{IEEEtran} 

\usepackage[utf8]{inputenc}
\IEEEoverridecommandlockouts                              

\usepackage[dvipsnames]{xcolor}
\usepackage{tikz}
\usepackage{graphicx}
\graphicspath{{figures/}} 
\usepackage{dcolumn}
\usepackage{bm}

\usepackage{svg}
\usepackage{comment}
\usepackage{amssymb}
\usepackage{amsthm}
\usepackage{svg}
\usepackage{amsmath,siunitx,booktabs}
\usepackage{epstopdf}
\usepackage{cite}
\usepackage{titlesec}
\usepackage{bm}
\usepackage{nomencl}

\makenomenclature

\title{\LARGE \bf
Uncertainty estimation of pedestrian future trajectory \\using Bayesian approximation 
}

\author{Anshul Nayak$^{1}$,  Azim Eskandarian$^{2}$, Zachary Doerzaph$^{3}$ 
\thanks{*This work was supported by SAFE-D}
\thanks{$^{1}$Anshul Nayak is a Ph.D. student in Autonomous Systems and Intelligent Machines Lab,
        Virginia Tech, USA. 
        {\tt\small anshulnayak@vt.edu}}%
\thanks{$^{2}$Azim Eskandarian is the Department Head and Nicholas and Rebecca Des Champs Chaired Professor in Mechanical Engineering, 
Virginia Tech, USA
        {\tt\small eskandarian@vt.edu}}%
\thanks{$^{3}$Zachary Doerzaph is the Executive Director at Virginia Tech Transportation Institute (VTTI) and 
associate professor in Biomedical Engineering and Mechanics,
        Virginia Tech, USA.
        {\tt\small zdoerzap@vt.edu}}%
}
\begin{document}

\maketitle
\pagenumbering{arabic}

\begin{abstract}

Past research on pedestrian trajectory forecasting mainly focused on deterministic predictions  which  provide only point estimates of future states. These future estimates can help an autonomous vehicle  plan its  trajectory and avoid collision.  However, under dynamic traffic scenarios, planning based on deterministic predictions  is not trustworthy. Rather, estimating the uncertainty associated with the predicted states with certain level of confidence can lead to  robust path planning. Hence, the authors propose to quantify  this uncertainty during forecasting using stochastic approximation which  deterministic approaches fail to capture.  The current method is simple and applies Bayesian approximation during inference to standard neural network architectures for estimating  uncertainty. The authors compared the predictions between the probabilistic neural network (NN) models   with the  standard deterministic models.  The results indicate that the mean predicted path of probabilistic models was closer to the ground truth when compared with the deterministic prediction.  Further,  the  effect of  stochastic dropout of weights and long-term prediction on future state uncertainty has been studied. It was found that the probabilistic models  produced better performance metrics like average displacement error (ADE) and final displacement error (FDE). Finally, the study has been extended to multiple datasets  providing a comprehensive comparison for each model.
\vspace{0.2cm}


\end{abstract}

\nomenclature[01]{\(x, y\)}{x and y  position of pedestrian from dataset }

\nomenclature[02]{\(u, v\)}{Pedestrian velocity along x and y direction}

\nomenclature[03]{\(\hat{x}, \hat{y}\)}{predicted x and y position }

\nomenclature[04]{\(\mathrm{X},\mathrm{Y}\)}{Training data and training label respectively}

\nomenclature[05]{\({x^{*}}, {y^{*}}\)}{test data sample and predicted outcome}

\nomenclature[06]{\(\theta\)}{Weight parameter of the neural network}

\nomenclature[07]{\(\Sigma_{xx}\), \(\Sigma_{yy}\)}{ covariance matrix along x and y direction}

\nomenclature[08]{\(\sigma_{x}, \sigma_{y}\)}{ standard deviation along x and y direction}

\nomenclature[09]{\(\mu_{x}, \mu_{y}\)}{Mean along x and y direction}

\nomenclature[10]{\(\bar{y}^{*}\)}{ Mean predicted path of trajectory distribution }

\nomenclature[11]{\(\bar{\Sigma}_{y^{*}}\)}{ Variance of trajectory distribution }

\nomenclature[12]{\(T_{f}\)}{ Forward prediction horizon (s) }

\nomenclature[13]{\(p\)}{stochastic dropout probability }

\printnomenclature[1.75cm]

\section{INTRODUCTION}

\subsection{Motivation}

For a self-driving car, awareness of the surrounding environment is crucial for correct and safe maneuvering \cite{Zhang_ce1}\cite{Nayak}.
Especially,  complex maneuvers require a trustworthy estimate of future states of vulnerable road users (VRUs) like pedestrians and bicyclists \cite{Eskandarian}. Continuous  progress has been made towards  predicting the motion of the vulnerable users  with a certain degree of effectiveness \cite{Xue}\cite{Hao}. However, most prediction models are  deterministic and provide only point estimates  of the future states \cite{Houenou}. Such  assumptions  may be helpful for specific scenarios but in a dynamic environment with multiple interactions, deterministic   predictions can be inaccurate. Since, humans tend to change directions swiftly, deterministic predictions may fail to capture this randomness in pedestrian trajectory and thus ignoring the associated uncertainty with the motion.  A more robust  approach will be to provide a probability distribution based on the likelihood of  pedestrian's location for each predicted state rather than a single point estimate. The uncertainty associated with predicted states can enable autonomous vehicles achieve  uncertainty-aware motion planning that will be more robust and trustworthy compared to planning based on deterministic prediction. 

\begin{figure}[ht]
 \centering
   {\includegraphics[width = 0.35\textwidth ] {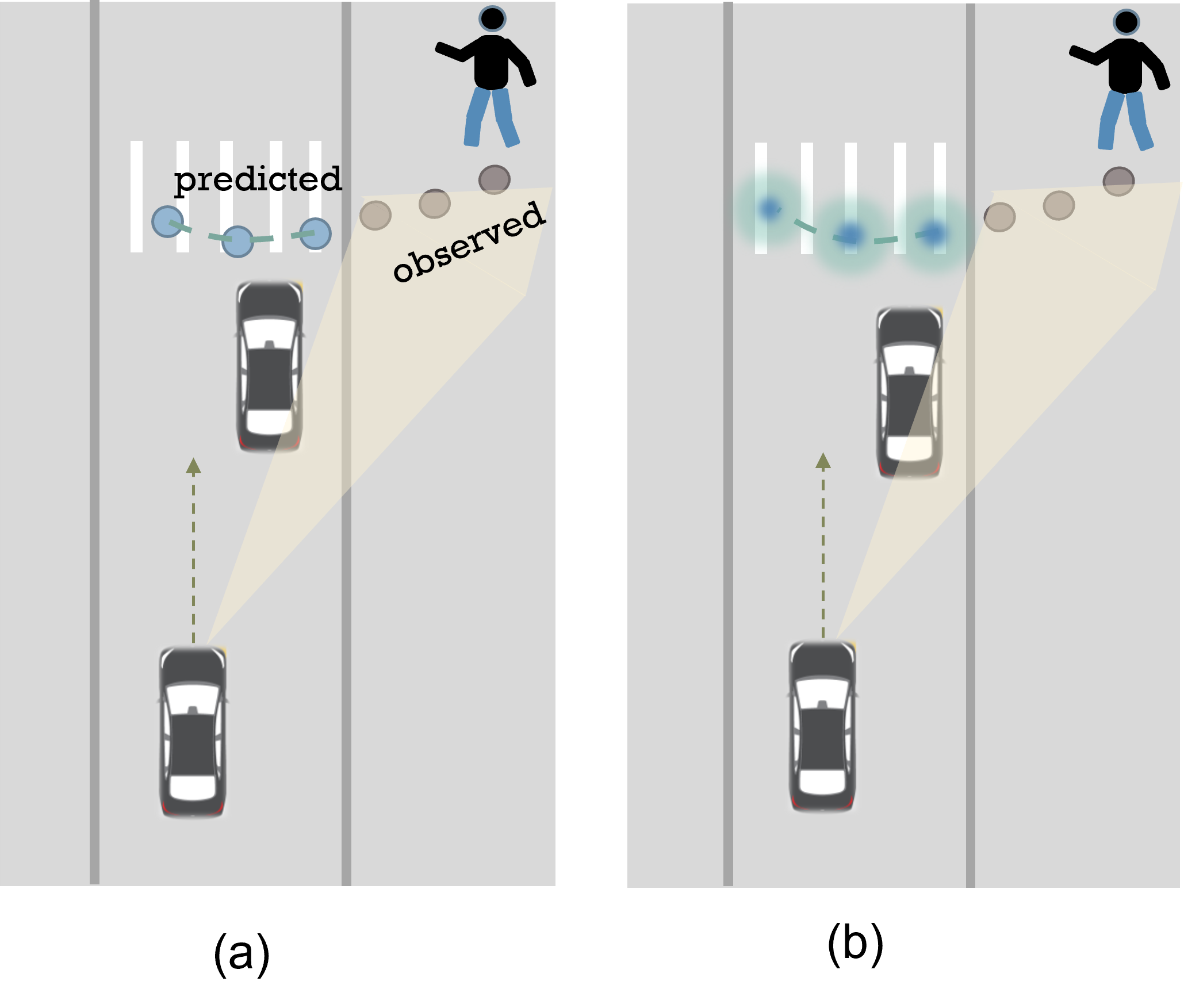}}
    \caption{(a) Deterministic trajectory forecasting with point prediction  (b) Probabilistic trajectory forecasting with associated uncertainty } \label{Road_scenario}
 \end{figure}

For instance, the deterministic prediction outputs point estimates of future states (Figure \ref{Road_scenario}a). However, during long-term forecasts or multiple actor interactions, the deterministic predictions deviate  from the ground truth. In such a scenario, the planning algorithm may be uncertain about the future states and the risk of collision will be  significantly high. Therefore, a robust and trust-worthy planning algorithm requires improvement in  confidence of the predicted states.  This can be achieved through probabilistic prediction of states with associated uncertainty (Figure \ref{Road_scenario}b). The risk-aware region around the future state captures all the probable locations the pedestrian can be present at a future time with certain confidence. It enables planning algorithms to either completely avoid the region or plan intelligently  to execute uncertainty-aware motion planning  that is more robust and  less prone to collision \cite{Abbeel}. Hence, the current-study proposes a risk-aware motion prediction model for pedestrians which can be further combined with intelligent planning algorithms. The current  model can perform long-term motion prediction with high positional accuracy. The uncertainty in motion is estimated in the form of a distribution of trajectories with mean and variance such that the ground truth mostly lies within 95\% of the confidence interval of the predictions.

\subsection{Related Work}
 
In recent years, Deep Neural networks (DNNs) have been used extensively for pedestrian trajectory forecasting. Most of the networks are based on recurrent neural network (RNN) architecture  which captures the temporal dynamics of  sequential data \cite{Suo}. Although,  RNNs should retain complete information of a temporal sequence, practically they fail to propagate long-term dependencies. Hence, Long short-term memory, LSTMs \cite{Park}\cite{Florent} have been used for sequential prediction due  to its improved capability in back propagating long-term error. Seminal works like scene-LSTM \cite{Manh} and social-LSTM\cite{Alahi} have used LSTM network architecture to incorporate either scene information or social pooling between multiple pedestrians for enhanced trajectory forecasting. However, most of the prior work on LSTM focused on improving prediction accuracy and did not stress on quantifying uncertainty.
Recently, Convolutional Neural networks have been used for trajectory prediction. Especially, a fast convolutional neural network (CNN) based model compared a 1D convolutional model with LSTM and showed improvement in  temporal representation of trajectory  \cite{Nikhil}. Further, Simone et.al \cite{Simone} elaborated upon the previous work by introducing novel preprocessing and data augmentation techniques to outperform other complex models. However, both the models output  deterministic estimates of future state and do not quantify uncertainty.  The deterministic predictions may not be trustworthy in complex traffic scenarios. Hence, probabilistic inference of predictions can be useful in safety-critical tasks like collision avoidance \cite{Xihui} and uncertainty-aware motion planning.

 Traditionally, Kalman filter has been used for uncertainty estimation \cite{Ammoun} but it fails to capture non-linearities during long-term forecasts.  Deep learning methods like Gaussian processes \cite{Ellis} and Gaussian Mixture Model \cite{Wiest} have been used for probabilistic forecasting too. For instance, a GP model with unscented Kalman Filter (UKF) was used for long term prediction of obstacle for collision avoidance \cite{Sun}.  However,  Gaussian process kernels are infinite-dimensional and require extensive parameter-tuning for accurate prediction. Further, methods like stochastic reachability analysis  has been  performed by assigning  probabilities to future states in the reachable set \cite{Khattar}. However, reachability analysis \cite{Althoff} often estimates all possible future states resulting in a large infeasible set.  Recently, non-linear architectures like neural networks are used for probabilistic trajectory forecasting. A CNN based architecture was used to capture the uncertainty but for short-term forecasts only \cite{Zernetsch}. Further,  an LSTM architecture could predict long term  probabilistic estimates of future states using an occupancy grid \cite{Kim}.  The problem was formulated as a multi class classification problem using softmax to quantify probabilistic distribution of future states over the occupancy grid. However,   softmax  often leads to over confident prediction and the model can be uncertain with high softmax output \cite{Gal}.  More sophisticated models like the Bayesian Neural networks (BNN) \cite{Zhang} have been used recently to capture the uncertainty in time series.

 Typically, a BNN uses a distribution over prior  to formulate a  posterior distribution which is used  to quantify uncertainty during prediction. It can capture three types of uncertainties, model uncertainty termed as epistemic uncertainty, inherent noise in data also known as aleatoric uncertainty, and model misspecification that occurs when testing data is different from the training dataset. In this paper, our main focus is to predict epistemic uncertainty associated with future trajectories. Although, Bayesian network accurately captures uncertainty, the inference becomes challenging due to a large number of model parameters. This often requires the incorporation of variational methods for Bayesian inference that can reduce computational costs. Recently, Monte Carlo (MC) dropout has been used as a variational method  for uncertainty estimation in time series forecasting \cite{Zhu} without any significant change to the network architecture. 
 
 In the current work, we have elaborated upon this Bayesian approximation of using MC dropout for pedestrian uncertainty estimation.
 Our work primarily focuses on comparing  the prediction performance of deterministic and probabilistic models. We have quantified and compared the uncertainty during trajectory forecasting using  three popular neural network architectures for time-series forecasting namely LSTM, 1D CNN and CNN-LSTM. Our novelty lies in showing   the importance of probabilistic forecasting of future states over deterministic  predictions and also providing a detailed performance comparison between each probabilistic model with its deterministic  model.  To our knowledge, our work is also novel in the sense that we also show the effect of  long-term forecasts   as well as stochastic dropout of weights on  uncertainty of future trajectory. Moreover,  this work also provides a comprehensive performance comparison of both probabilistic and deterministic models on popular pedestrian datasets.

 The remainder of the paper is organized as follows. In section \ref{methods}, we introduce the methods for probabilistic forecasting with Bayesian Neural network followed by describing  each neural network architecture;  the encoder-decoder model, the convolutional model and the CNN-LSTM model  along with Monte Carlo dropout. In section \ref{experiments}, we describe  the data preprocessing, implementation details and performance metrics.  In section \ref{results}, we discuss the results of  uncertainty quantification and provide a comprehensive  study on  the effect of future prediction horizon and stochastic dropout on performance metrics.  Section \ref{conclusion} concludes our study.

\section{METHODS}\label{methods}

\begin{figure}[ht]
 \centering
   {\includegraphics[width = 0.48\textwidth ]{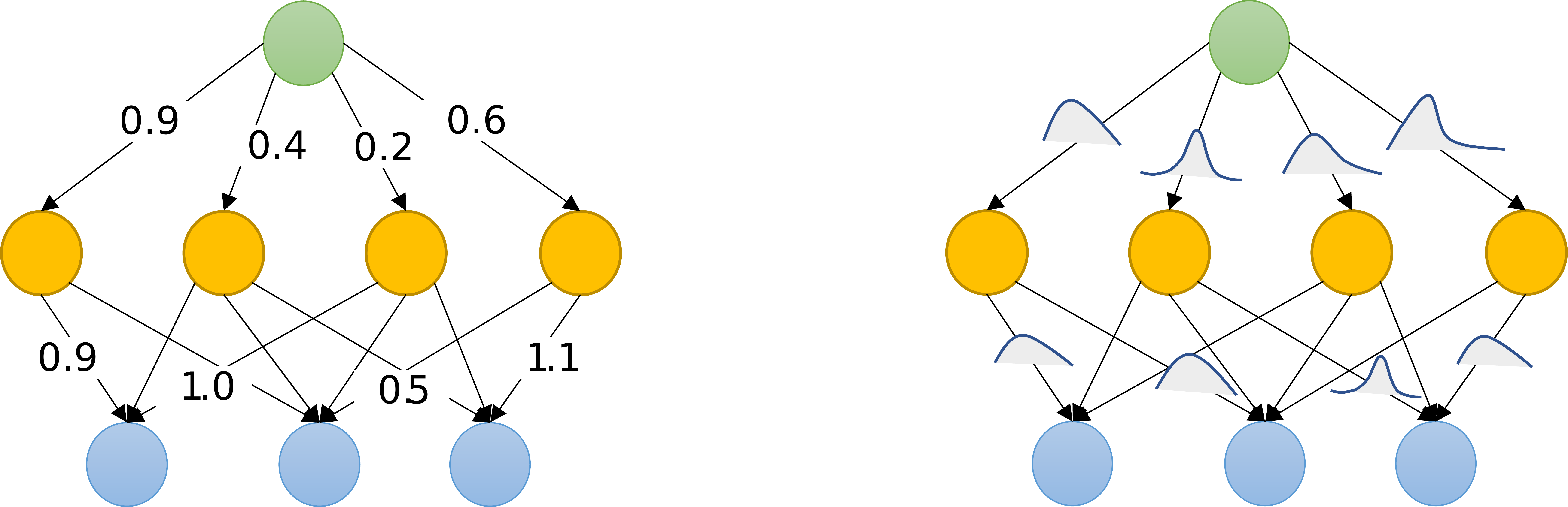}}
    \caption{(a) Deterministic Neural network with point weights (b) Stochastic Neural Network with weight distribution } \label{BNN_diagram}
 \end{figure}
  
  \begin{figure*}[ht]
 \centering
   {\includegraphics[width = 0.9\textwidth ] {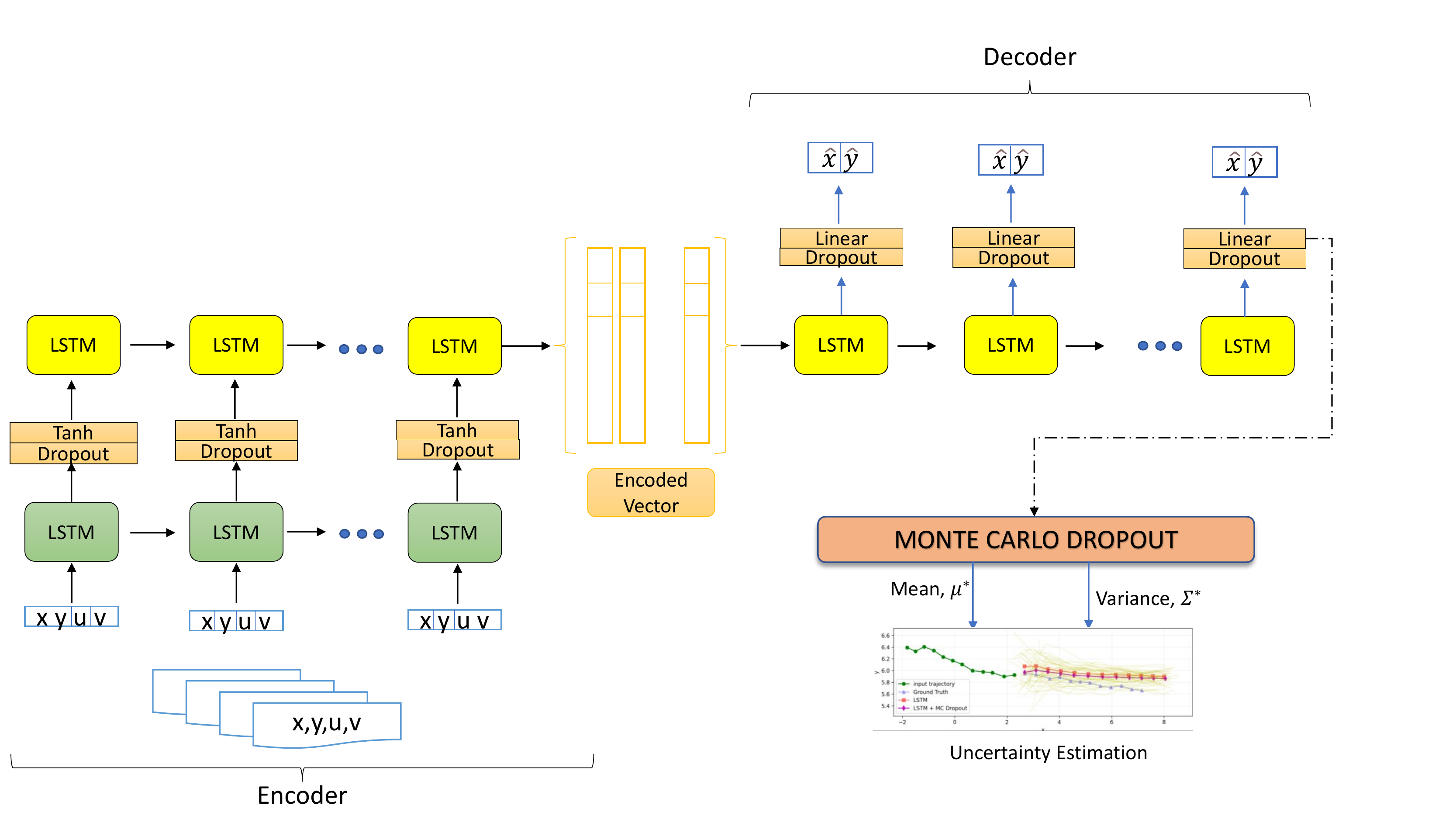}}
    \caption{LSTM architecture with Monte Carlo Dropout} \label{LSTM architecture}
 \end{figure*}
Deep learning architectures have been used extensively for prediction tasks. However, most networks are deterministic  generating  point estimates without any confidence interval (Figure \ref{BNN_diagram}a). Conversely, a Bayesian Neural Network (BNN) produces uncertainty-aware predictions  based on a stochastic network.  Although, stochastic networks might provide better point estimates than other standard neural networks, the main aim of stochastic networks is to provide trustworthy uncertainty estimates for predictions. Typically, a stochastic neural network introduces stochastic weights or activation functions (Figure \ref{BNN_diagram}b) into the model and is trained using Bayesian approach  as:

\begin{equation}
    {P}(H|D) = \frac{{P}(D|H)\,{P}(H)}{ {P}(D)}
\end{equation}
${P(H|D)}$ is the posterior which implies the hypothesis, H is statistically updated using inference based on data, D.  It is the main aspect of a stochastic network as it captures the model uncertainty known as epistemic uncertainty. The probability ${P}(D|H)$ is called likelihood which captures the inherent noise in the data also known as aleatoric uncertainty. Meanwhile, ${P}(H)$ is  the prior while ${P}(D)$ is the evidence.  The prior samples  from a stochastic distribution of weight with probability, ${P}(H)$ unlike standard NNs which have deterministic weights  (Figure {\ref{BNN_diagram}}). As a result, the posterior  distribution also becomes  stochastic and the metrics can be estimated by computing the mean of predicted distribution alongwith its associated variance. Particularly, stochastic networks such as  Bayesian neural networks can be used for time-series forecasting with applications expanding to risk-aware prediction of vulnerable road users. A BNN has several advantages over traditional neural networks such as estimating uncertainty, classifying the uncertainty into epistemic and aleatoric uncertainty as well as integrating the  knowledge of prior into the model. For instance, provided a set of training input states, X = {$x_{1}, x_{2}, ...,x_{T}$} and outputs, Y = {$y_{1}, y_{2}, ...,y_{T}$} for a pedestrian, the  distribution of its future predicted states, $y^{*}$ can be computed by marginalizing the posterior, ${p}(\theta|X,Y)$ over some new input data point, $x^{*}$:

\begin{equation}
    {p}(y^{*}|x^{*},X,Y) = { \int_{\theta}^{} {p}(y^{*}|x^{*},\theta'){p}(\theta'|X,Y) \,d\theta' }
\end{equation}
$\theta'$ represents weights and  $p(\theta')$ refers to the probability of sampling from prior weight distribution. During Bayesian prediction, computing the posterior, ${p}(\theta'|X,Y)$ can be quite challenging and many sampling algorithms like  Monte Carlo Markov Chain (MCMC) and variational inference \cite{Jospin} have been used to approximate the posterior. But, most of the sampling methods are either computationally expensive or introduce a lot of parameters into the model which stems into a complex  problem.
 Further, a BNN incurs additional computation costs due to its  non-linearity.  Hence, we use the  Monte Carlo (MC) dropout method \cite{Gal}  which has been used to accurately approximate a  Bayesian neural network without  significantly changing the model. The MC dropout method assumes that weights are stochastically dropped during inference. This process is then repeated for N forward passes to generate a random distribution of predicted values.   Therefore, we have  applied Monte Carlo dropout to different neural network architectures for uncertainty quantification during predictions. 

\subsection{Encoder Decoder Model}

We developed an encoder-decoder based simple LSTM architecture  to predict future trajectory of pedestrians up to multiple time horizons. An encoder creates an embedding of
essential features as a series of encoded space vectors which
the decoder uses for estimating outputs. Suppose,  $\{{x_{t}\}}_{T}$ represents the x-position up to T time steps as $\{x_{1}, x_{2}, ..,x_{T}\}$, then, the encoder embeds the data into an encoded space using non-linear function as $e = g(x)$. The decoder then uses the encoded features to construct $F$ forward time steps, $\{x_{T+1}, x_{T+2},...,x_{T+F}\}$.  The current architecture has two LSTM cells for the encoder and one for the decoder  (Figure \ref{LSTM architecture}).  Through an ablation study, we found encoding both position and velocity  of the pedestrian was beneficial for accurate prediction rather than encoding only position information. Hence, the input  data  for LSTM layers is a multivariate time series with four features $\{x,y,u,v\}$ corresponding to x and y position and velocity respectively. The output of the neural network has similar features and the  predicted $\hat{x}$,  $\hat{y}$  show pedestrian's future position. For regularization,  dropout of weights with a certain probability, p  followed by \textbf{tanh} activation has been implemented. The steps are repeated for each LSTM layer to create the encoded vector space which carries the essential features of the training data. 
The decoder then uses the latent encoded vector to predict the future motion. The decoder architecture has a single LSTM layer that takes the encoded vector space as input. The LSTM layer is then followed by a  dropout layer with linear activation to estimate the output.  

\vspace{0.25cm}
\subsection{Convolutional Model}

\begin{figure}[h!]
 \centering
   {\includegraphics[width = 0.5\textwidth ] {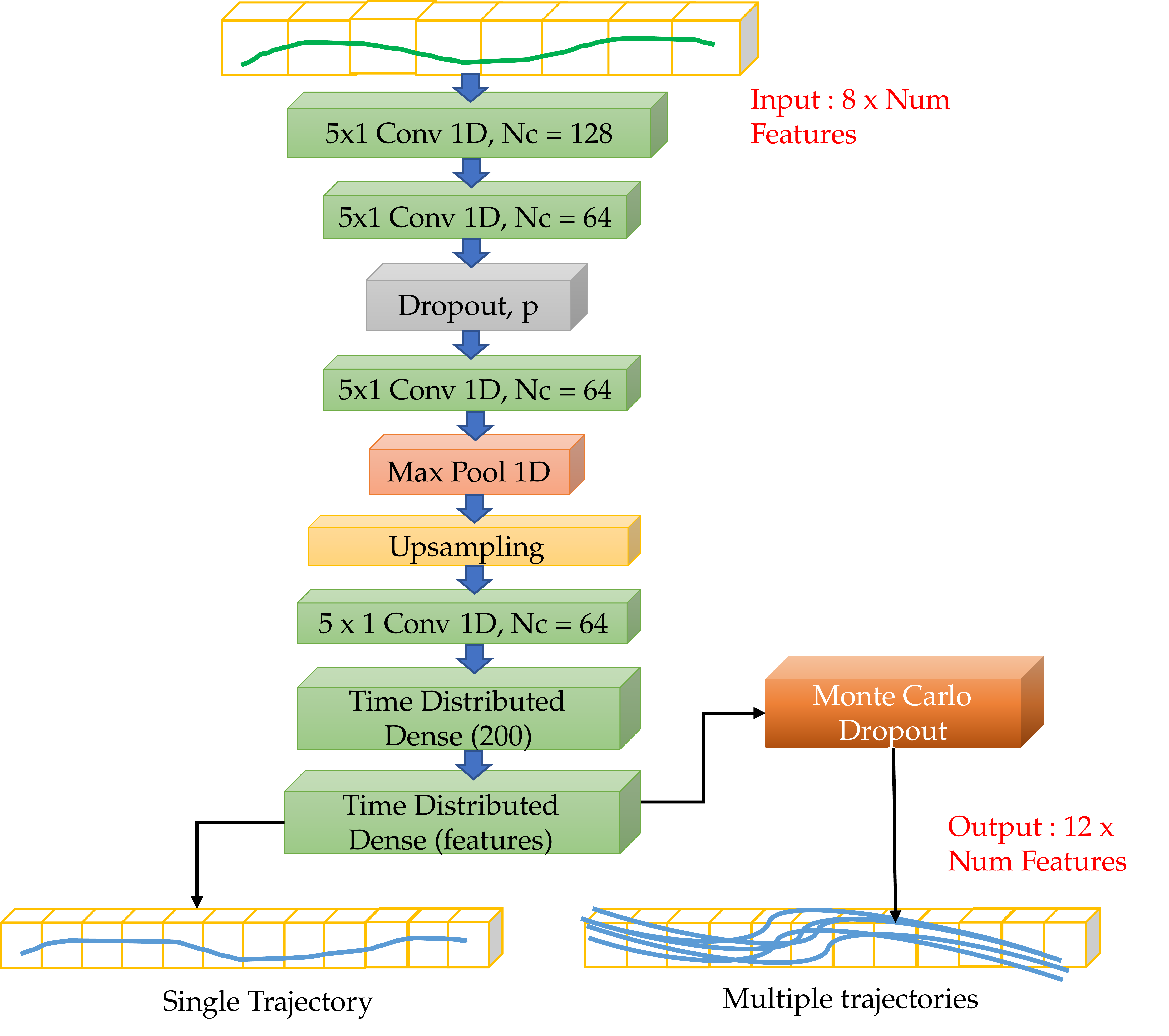}}
    \caption{CNN architecture with Monte Carlo Dropout} \label{CNN architecture}
 \end{figure}

Recently, convolutional networks like 1D CNNs have been used for time series analysis. The CNN model is a simple sequence-to-sequence architecture that uses  convolutional layers to handle temporal representations.  In the current 1D CNN model, we represent the past trajectory as a one-dimensional channel with four features, $\{x,y,u,v\}$. We constantly pad the input at each convolutional layer so that the number of features in both input and output remains same. We have used "causal" padding for modeling temporal data which eliminates the curse of auto  correlation such that the output[t] does not depend upon input[t+1].  In total, three 1D Convolutional layers with 128, 64, 64  filters respectively have been used to extract features (Figure \ref{CNN architecture}). We use a kernel  size of 5 which showed better root mean-squared error (RMSE) in an ablation study with other odd  kernel sizes \{3,7\}.   A single dropout layer with probability, p is applied after the first two convolutional layers to prevent overfitting.  Further, 'ReLU' activation function has been applied to all the convolutional layers. We observed a 1D Max Pool layer followed by upsampling performed better than global average pooling. Finally, for many-to-many predictions, time distributed dense layer has been used  to predict multiple time steps simultaneously generating a single trajectory. Further,  MC dropout can be applied during inference to generate a distribution of  trajectories.    

 \vspace{0.25cm}
\subsection{CNN-LSTM}
A CNN-LSTM architecture is a hybrid network that uses both  convolution and LSTM layers  as an end-to-end model for time-series forecasting.  Unlike the LSTM encoder-decoder model,   one-dimensional convolutional layers are used for feature extraction to create an embedding rather than LSTM layers. The current model has two one-dimensional convolution layers  with filters 128 and 64 respectively. Each convolution layer is followed by a dropout layer where weights are dropped with a certain probability, p to prevent overfitting. In order to process the data into the format required by the LSTM, a Flatten layer is connected after convolution. Further, the  LSTM layer acts as decoder and utilises the features for prediction. Similar to previous models, a time-distributed layer at the end enables multi-step forecasting.

\vspace{0.25cm}
\subsection{MC Dropout}
Epistemic uncertainty during forecasting problems can be  estimated  from the mean and variance of the marginal distribution, ${p}(y^{*}|x^{*},X,Y)$. Recall that X and Y represent the training input and output samples while $y^{*}$ represents the predicted states for some new test sample, ${x^{*}}$.  The marginal distribution is  sampled from posterior, ${p}(\theta|X,Y)$ based on  test input, $x^{*}$.  The sampling process  can be challenging and computationally expensive due to non-linearity in BNNs. Therefore,  Monte Carlo dropout can be used as a variational approximation to such Bayesian Neural Networks \cite{Gal} without modifying the existing network architecture much.  Hence, model uncertainty  known as epistemic uncertainty can be estimated with ease without any additional computational cost, unlike other inference methods.

As discussed,  Monte Carlo dropout (MC dropout) has been applied to each network architecture during inference to quantify uncertainty (Figure \ref{LSTM architecture},\ref{CNN architecture}). The MC dropout model applies stochastic  dropout during test time to the neural network. Thus, for any new set of input, ${x}^{*}$, we compute the inference by random dropout at each layer of the model. The probability of dropout is set as p and the inference model is run N times to obtain a set of outputs $\{y_{1}^{*}, y_{2}^{*}, ...,y_{N}^{*}\}$. We can then estimate the mean, $\bar{{y}}^{*}$ and variance, $\bar{\Sigma}_{{y^{*}}}$ of the marginal distribution where the variance indicates  model uncertainty (Equation \ref{Performance_metrics}).

\begin{align}
\begin{split}
    \text{Mean,}\,\bar{{y}}^{*} &= \frac{1}{N}\sum_{n=1}^{N}{y^{*}}_{(n)} \\
     {{\text{Var}},}\,\bar{\Sigma}_{{y^{*}}} &= \frac{1}{N}\sum_{n=1}^{N} ({y^{*}}_{(n)} - \bar{{y}}^{*})^2
\end{split}
\label{Performance_metrics}
\end{align}

\vspace{0.25 cm}

\section{EXPERIMENTS}\label{experiments}

In this section, we discuss  the datasets, data augmentation,  implementation details for each network and the performance metrics.  Following common practice from literature \cite{Alahi}, we trained our models on publicly available pedestrian datasets. Two most popular datasets are the ETH dataset \cite{Pelligrini} which contains the ETH and HOTEL scene while the UCY dataset \cite{UCY} which contains the UNIV, ZARA1 and ZARA2 scenes.    In order to draw parallelism with past works \cite{Nikhil}, we studied 8 (3.2 secs) historical steps to predict 12 (4.8 secs) steps into the future. Further, we extended our study to predict multiple time horizons with long-term forecasts  upto 8 seconds into future. 

\subsection{Data Augmentation}
Initially, we trained our model  on the ETH  dataset only which contains approximately 420 pedestrian trajectories under varied crowd settings. However, a small number of trajectories is insufficient for training. Therefore, we performed  data augmentation using Taken's Embedding theorem \cite{Takens}. We used a sliding window of T = 1 step to generate multiple small trajectories out of a single large trajectory. For instance, a  pedestrian's  trajectory of 29 steps will result in  10 small $\{x,y\}$ trajectory pairs of 20 steps each if past trajectory information of  8 steps is used for predicting  12 steps into future. In total, we constructed 1597 multivariate time series sequences which we split into 1260 training  and 337 testing sequences for the ETH hotel dataset.   Further, in Section \ref{datasets}, we train on the ZARA1, ZARA2  datasets too  to provide a comprehensive comparison of uncertainty quantification across all the models.  

\subsection{Implementation details}
All the  neural networks are trained end-to-end using Tensorflow. Adam optimizer with  a learning rate of $1e-3$ was used  to compute the  mean-squared error (MSE) loss.  Each model was trained for 100 epochs with a batch size of 32. The LSTM encoder-decoder model was trained with 'tanh' activation while the other two models had 'ReLU' activation. 10\% of the training data was used for validation. The MSE error was monitored on the validation loss with callback functions like  EarlyStopping and ReduceLRonPlateau.  The model was compiled and fit using train and test data.  

\subsection{Performance Metrics}
The trained model is then used to predict the future position of the pedestrian. By default, the model predicts deterministic future states. However, probabilistic predictions can be inferred using Monte Carlo dropout. We can run the stochastic inference using MC dropout repeatedly to generate a distribution of trajectories (Figure \ref{LSTM + MC Dropout}).  The mean of the distribution represents the predicted path while the associated  variance quantifies uncertainty.  We adopt the widely used performance metrics \cite{Alahi} namely average displacement error (ADE) and final displacement error (FDE)  for prediction comparison between the deterministic and probabilistic models. \hfill
\vspace{0.1cm}

(a) Average Displacement Error (ADE):  It refers to the mean of Euclidean distance over all estimated points of every trajectory with the corresponding points in ground truth. 

\begin{align}\label{ADE}
\begin{split}
    \text{ADE} &= \frac{1}{T}\sum_{t=t_{0}}^{t_{f}}
    ||{\hat{Y}_{(t)}- {Y}_{(t)}}|| 
    \end{split}
\end{align}

(b) Final Displacement Error (FDE): The mean of Euclidean distance between the predicted final destination and true final destination across all trajectories.

\begin{align}\label{FDE}
\begin{split}
    \text{FDE} &= 
    ||{\hat{Y}_{(t_{f})}- {Y}_{(t_{f})}}|| 
\end{split}
\end{align}

where $\hat{Y}_{t}$ is the predicted location at timestamp t and $Y_{t}$ is the ground truth position.

\section{Results and Discussion}\label{results}
In this section, we discuss about the uncertainty estimation and performance metrics of pedestrian trajectory prediction. Initially, we quantify uncertainty in prediction using the probabilistic   models  based on the ETH dataset. We also provide a confidence estimation of our predictions with respect to the ground truth. Later, we have have shown the effect of varying forecast horizon and dropout on performance metrics for all the deterministic and probabilistic models. Finally, we provide a comprehensive comparison of performance metrics among all the models across multiple datasets.

\subsection{Uncertainty Estimation}

\begin{figure}[h!]
 \centering
   \includegraphics[width = 0.425\textwidth,  trim = 1cm 3.75cm 0cm 4.5cm, clip ] {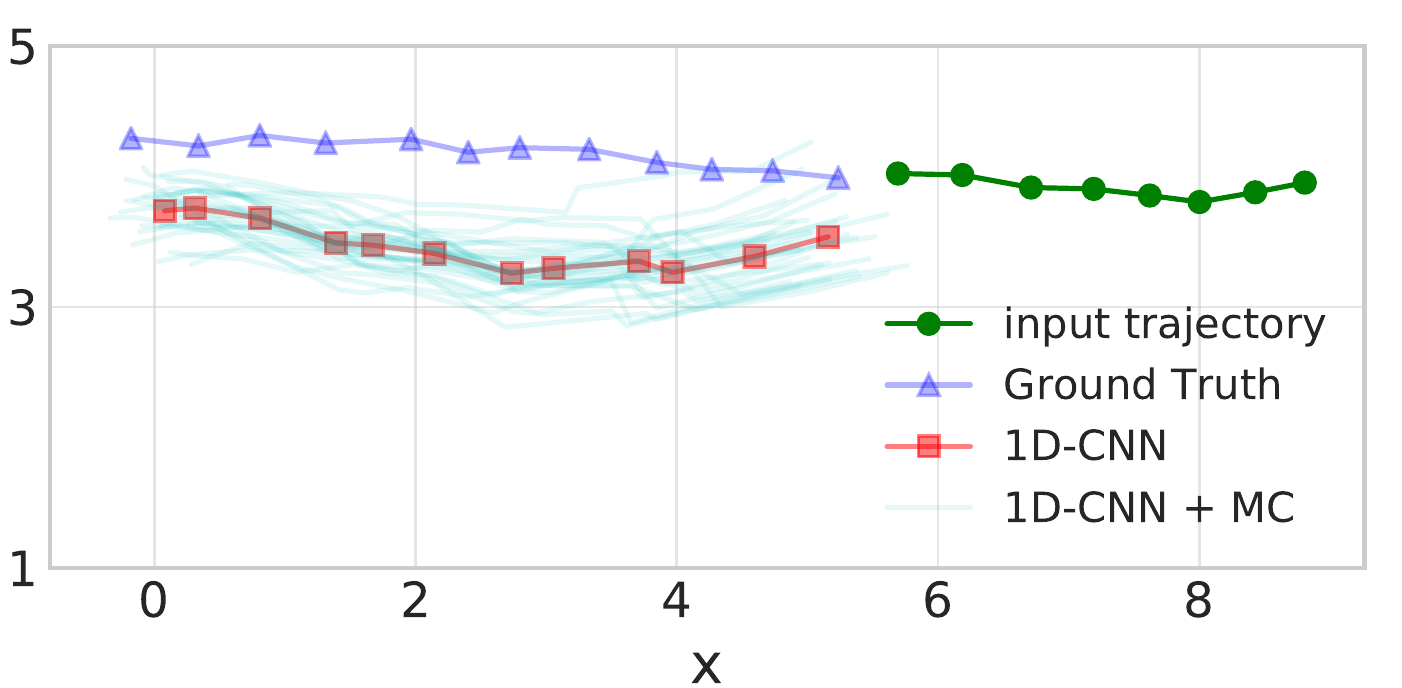}
    \caption{Trajectory prediction using Monte Carlo dropout with the 1D CNN network.} \label{LSTM + MC Dropout}
 \end{figure}

For predicting uncertainty and evaluating its trustworthiness, we generated a distribution of trajectories using MC dropout. We have shown the uncertainty estimation of a single pedestrian trajectory from the ETH dataset using the  1D CNN network (Figure \ref{LSTM + MC Dropout}). For the current scenario, we predict  12 steps or 4.8 seconds into the future  based on  3.2 seconds of past trajectory data.
A single trajectory (red) is generated by deterministic prediction and provides point estimates of future states.  However, we generate probabilistic predictions by applying MC dropout to each neural network model. For instance, the 1D CNN with MC dropout model was sampled  N = 30 times during inference with a stochastic dropout, p = 0.2 (20\% of the weights are randomly dropped)  to generate a distribution of N different trajectories. 
The  mean and variance of the distribution quantifies the uncertainty during trajectory prediction.

 \begin{figure}[h!]
 \centering
   \includegraphics[width = 0.45\textwidth,  trim = 0cm 1.0cm 0cm 2cm, clip ] {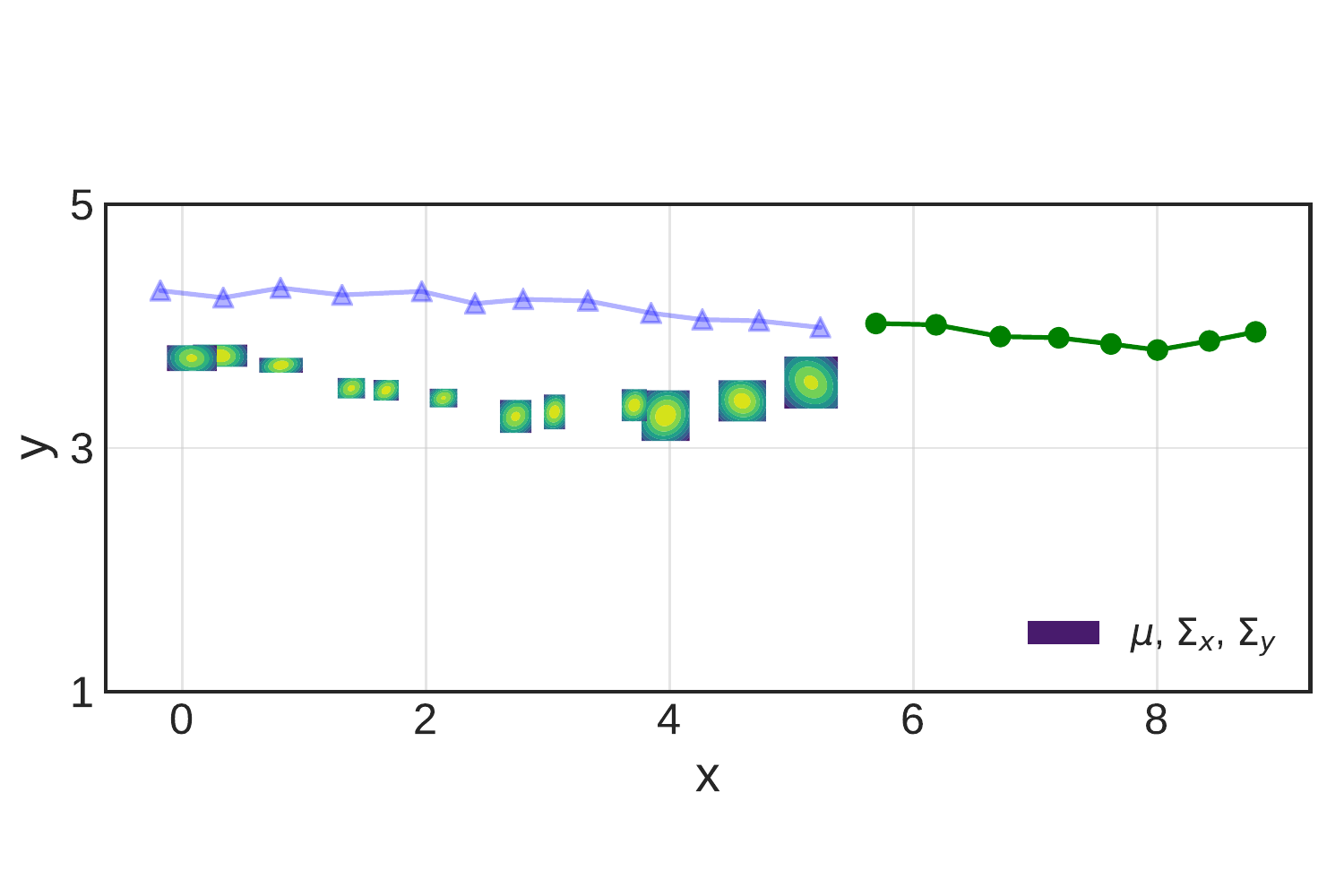}
    \caption{Bivariate Gaussian distribution of future state uncertainty } \label{Gaussian}
 \end{figure}

\begin{figure}[h!]
 \centering
   \includegraphics[width = 0.3\textwidth ] {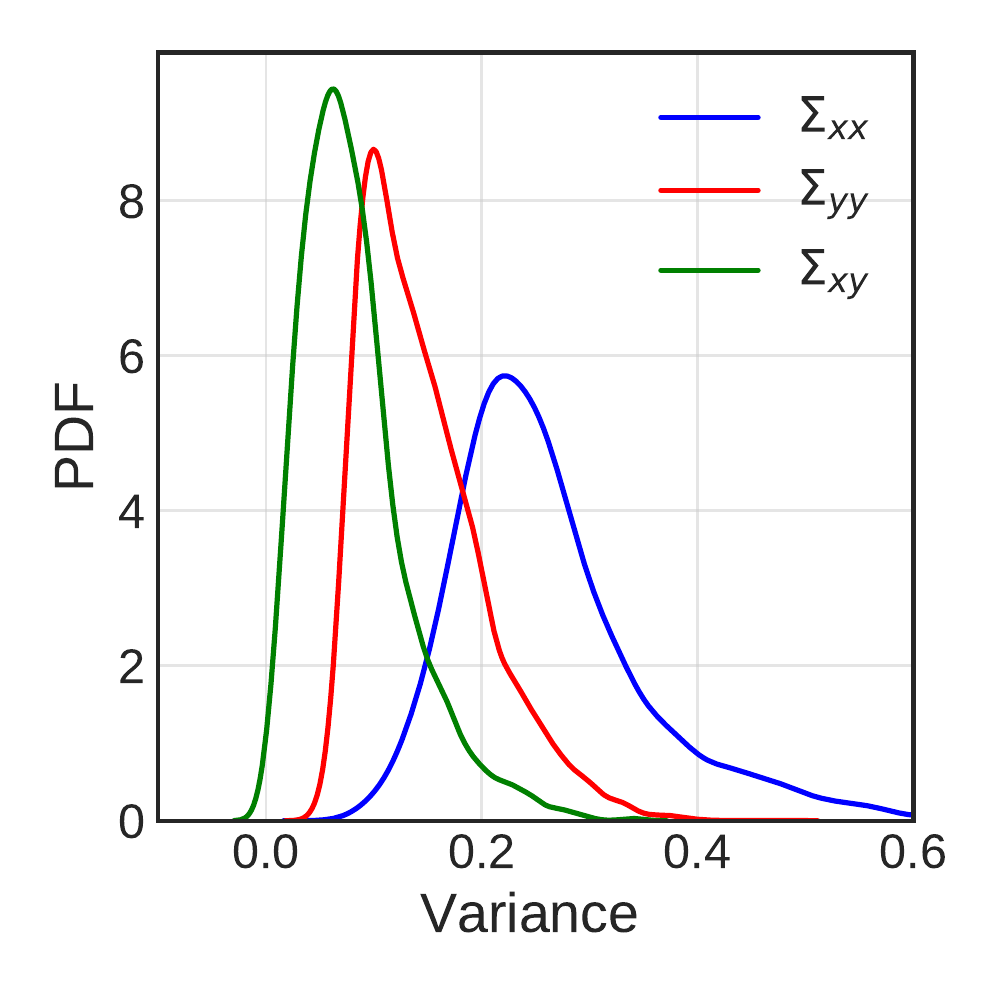}
    \caption{Variance in pedestrian motion along x, y and xy. } \label{Covariance Matrix}
 \end{figure}

The predicted trajectory distribution (Figure \ref{LSTM + MC Dropout})  shows the pedestrian's motion  along both x and y direction with predominant motion along x. Therefore, we need to quantify the mean and variance along both the directions and  treat the predicted trajectory cluster as a bivariate distribution.  At each prediction step,  the trajectories can be represented as  a cluster of N points distributed on the x-y domain.  Assuming  Gaussian distribution for each point cluster,  we can then estimate the mean and covariance showing the associated uncertainty at each step (Figure \ref{Gaussian}). The Gaussian representation of the predicted states thus shows the future states with associated uncertainty.  Further, we can compute the mean and covariance of this bivariate Gaussian distribution as:

\begin{align}
\begin{split}
    \begin{bmatrix} 
	{x}\\
	{y} 
	\end{bmatrix}
	 \sim \mathcal{N}
	  \begin{pmatrix} \begin{bmatrix} 
	\mu_{{x}}\\
	\mu_{{y}} 
	\end{bmatrix} 
	,
	   \begin{bmatrix} 
	\Sigma_{{x}{x}} & \Sigma_{{x}{y}} \\
		\Sigma_{{y}{x}} & \Sigma_{{y}{y}} \\
	\end{bmatrix}
	\end{pmatrix}
\end{split}
\vspace{0.2cm}
\end{align}
Here, \{$\mu_{{x}}$, $\mu_{{y}}$\} and \{$\Sigma_{{x}{x}}$ , $\Sigma_{{x}{y}}$, $\Sigma_{{y}{y}}$\}  represent the mean  and covariance  of pedestrian's movement along x-y domain respectively. Each covariance term shows the correlation of motion along one direction with respect to another. 
Our results  indicate that the covariance,  $\Sigma_{xx}$ is significantly higher compared to covariance along other directions (Figure \ref{Covariance Matrix}). Similar observation was seen across all the trajectories of ETH dataset. This shows that uncertainty in motion is more along x followed by y and xy. As the covariance $\Sigma_{xy}$ is negligible, only  the standard deviation, $\sigma = \sqrt{\Sigma}$ along x and y  were considered to quantify the uncertainty during trajectory prediction. 

\begin{figure}[h!]
  \centering
  \begin{tabular}{@{}c@{}}
    \includegraphics[width=.8\linewidth, trim = 1cm 3.65cm 0cm 4cm, clip]{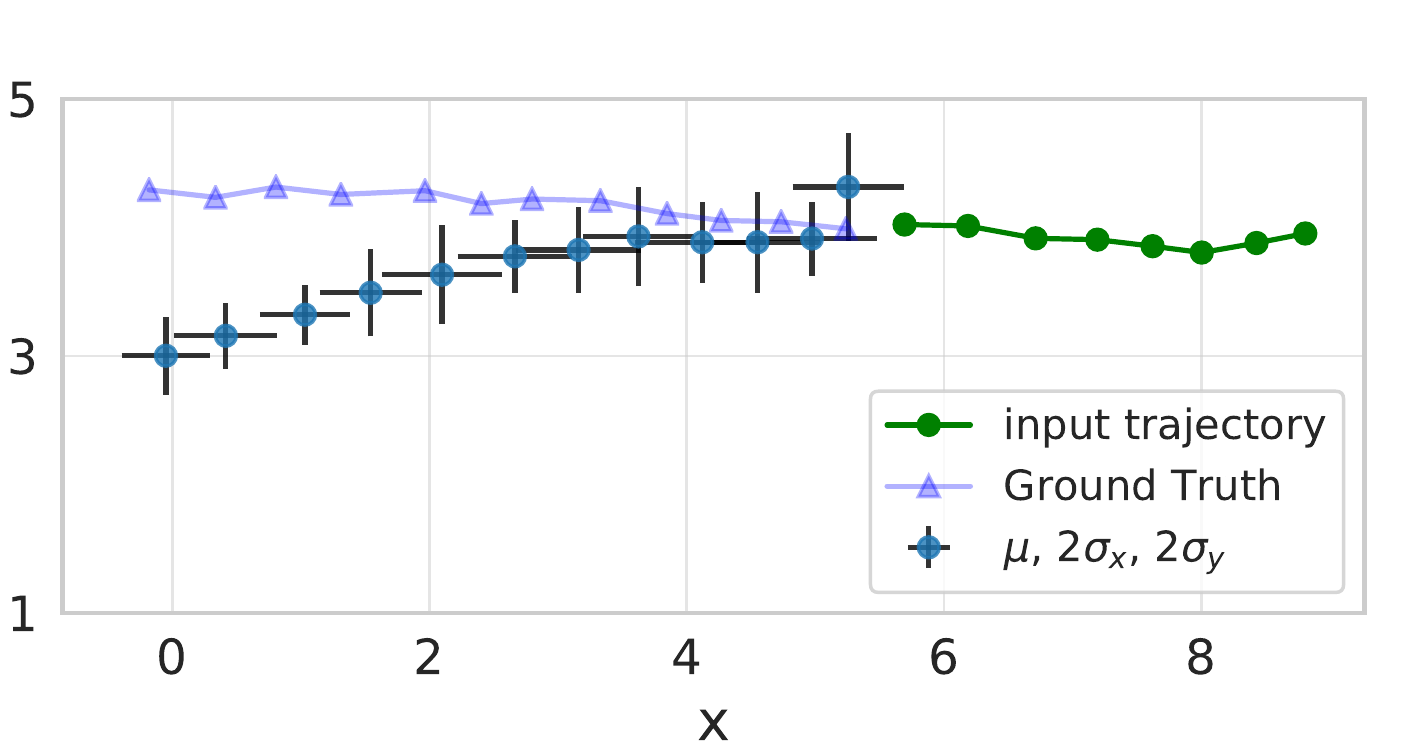} \\[\abovecaptionskip]
    \small (a) LSTM
  \end{tabular}
  
    \begin{tabular}{@{}c@{}}
    \includegraphics[width=.8\linewidth, trim = 1cm 3.65cm 0cm 3.75cm, clip]{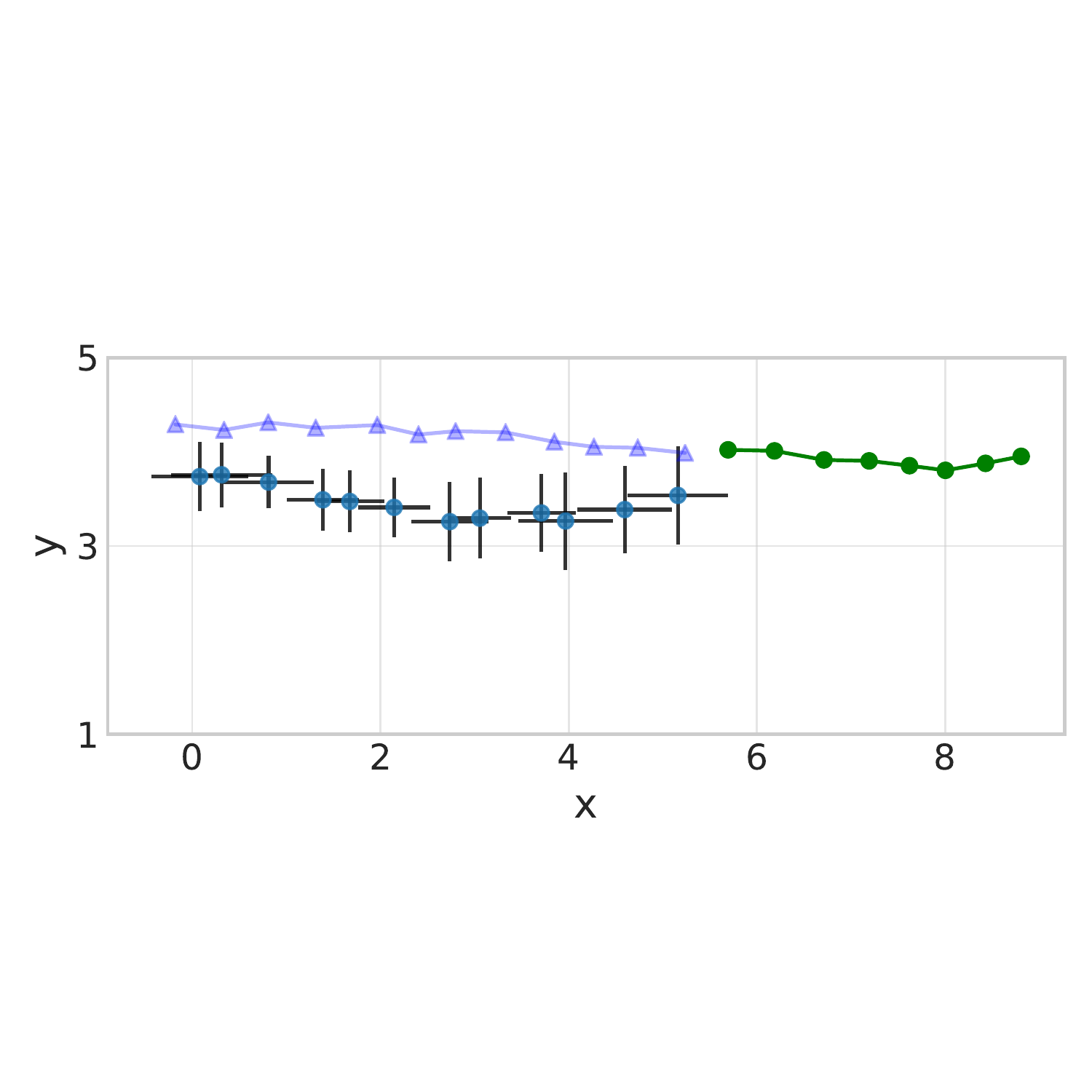} \\[\abovecaptionskip]
    \small (b)1D-CNN
  \end{tabular}

  \begin{tabular}{@{}c@{}}
    \includegraphics[width=.8\linewidth, trim = 1cm 3.65cm 0cm 3.75cm, clip]{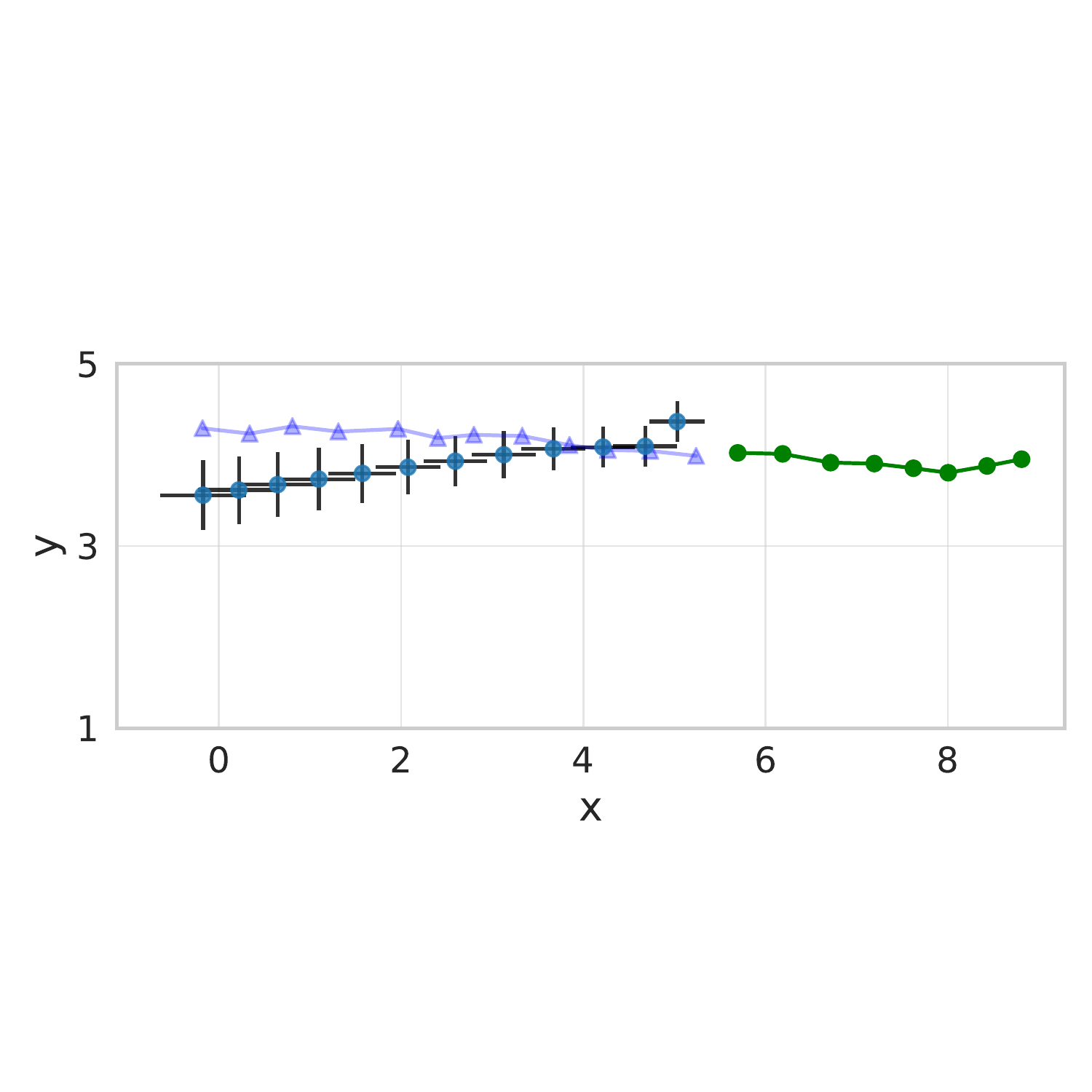} \\[\abovecaptionskip]
    \small (c) CNN-LSTM
  \end{tabular}

  \caption{Uncertainty quantification of pedestrian motion}\label{Uncertainty quantification}
\end{figure}

We compared the estimated uncertainty during prediction of each probabilistic model  with the ground truth, {\color{blue} \transparent{0.5}$\blacktriangle$}  (Fig \ref{Uncertainty quantification}). We have shown the mean predicted path  with two standard deviation ($2\sigma$ )
along  x and y direction to quantify uncertainty. The model takes 8 input states ({\color{ForestGreen} \transparent{0.85}{$\bullet$}}, green dot) to predict 12 states into future.  One can visualise the mean of the predicted distribution ({\color{RoyalBlue} \transparent{0.75}{$\bullet$}}, blue dot) alongwith the  standard deviations ({\color{Black}$\boldsymbol{+}$}) along x and y  for a single trajectory chosen from the ETH dataset (Fig \ref{Uncertainty quantification}).   On inspection, it seems the uncertainty predicted by the CNN-LSTM  model grows with prediction horizon while both LSTM and 1D-CNN models provide conservative probabilistic estimates that neither decease nor increase with time. Further, it appears that the final displacement error (FDE) is minimum for 1D CNN while it is maximum for LSTM.  However, to obtain conclusive evidence on predictive accuracy of each model, we need to consider the performance metrics of all possible trajectories  as well as estimate whether the ground truth lies within the 95\% ($2\sigma$) confidence interval of our predictions. 
 
 \begin{figure}[h!]
  \centering
  \begin{tabular}{@{}c@{}}
    \includegraphics[width=.8\linewidth]{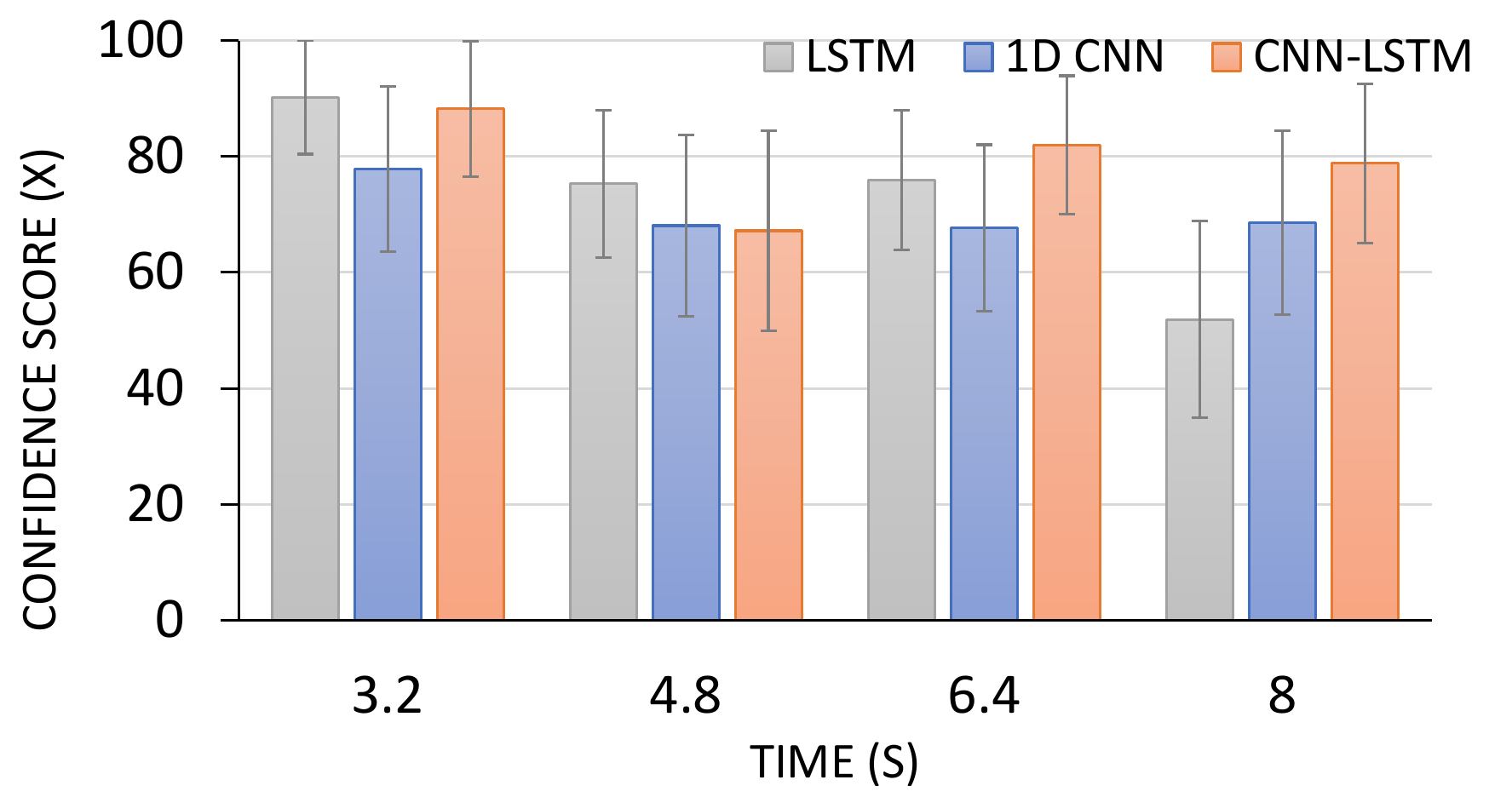} \\[\abovecaptionskip]
    \small (a) 
  \end{tabular}

  \begin{tabular}{@{}c@{}}
    \includegraphics[width=.8\linewidth]{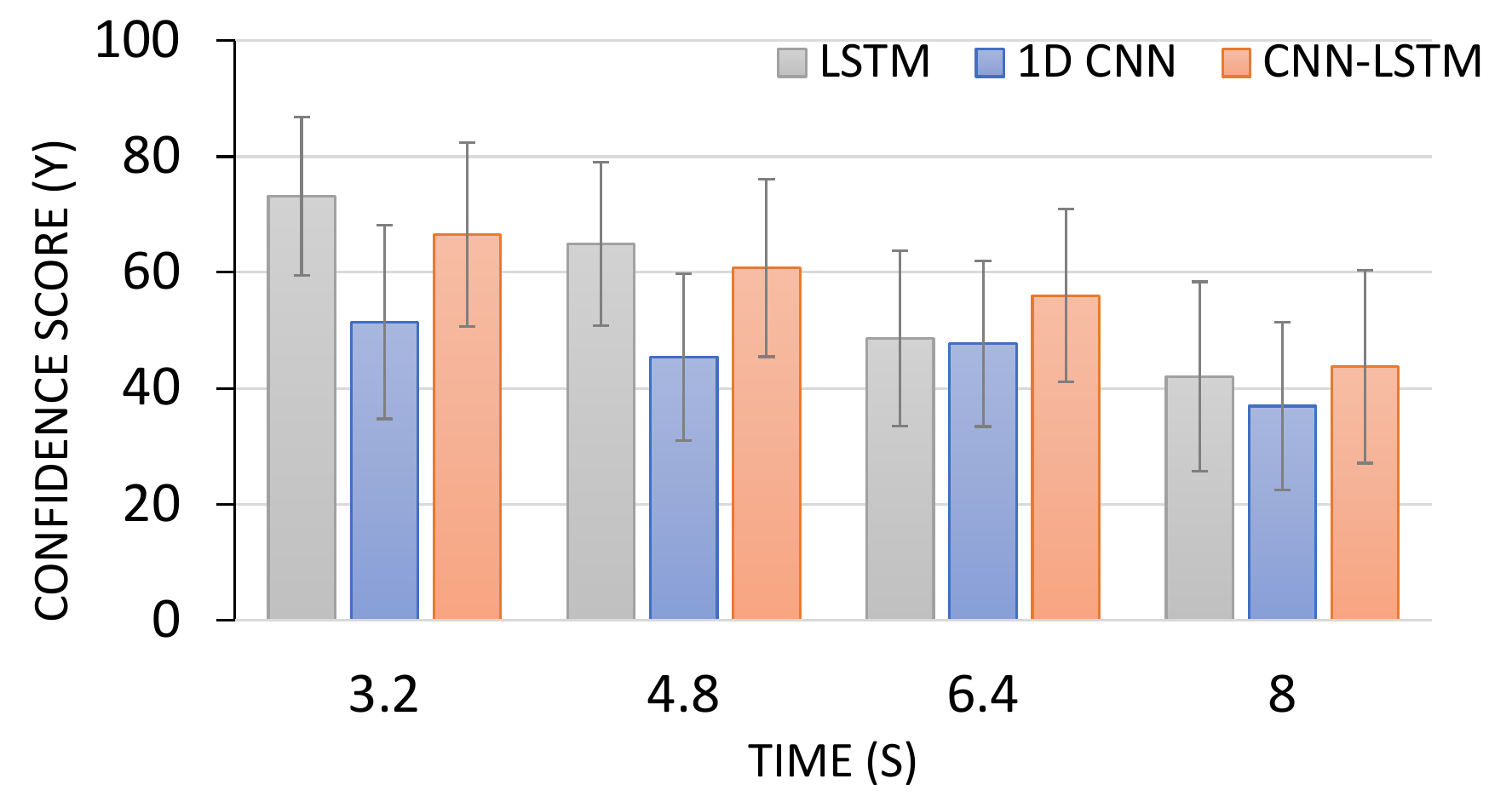} \\[\abovecaptionskip]
    \small (b)
  \end{tabular}

  \caption{Variation of dropout probability, p on (a) ADE and (b) FDE}\label{Confidence Interval}
\end{figure}

\vspace{0.1cm}
\subsubsection*{Confidence Interval}
\vspace{0.1cm}

 We define a parameter known as confidence score (\textit{CS}) to  check whether the ground truth $\{x_{true},y_{true}\}$  lie within two standard deviation ($2\sigma$) of our predicted distribution. 
 
 \begin{align}
{CS_{x}} &= \frac{|{x_{true} -\mu_{x}|_{i= 1,..F} < 2\sigma_{x}} }{F} \, \mathbf{x} \, 100  \nonumber \\
{CS_{y}} &= \frac{|{y_{true} -\mu_{y}|_{i= 1,..F} < 2\sigma_{y}} }{F} \, \mathbf{x} \, 100
\end{align}
\vfill
  Here, $F$ represents the number of predicted states into future. We predict the confidence score along x and y for each test trajectory and then take the mean across all trajectories to obtain a single confidence score for that prediction horizon. Further, we also show the variation of confidence score with prediction time horizon, $T_{F}$ = 3.2, 4.8, 6.4, 8 seconds into future.

The LSTM model with MC dropout (gray) provides a high mean confidence score over other models for forecasts upto 4.8 seconds. For instance at $T_{F}$ = 3.2 seconds, the LSTM model has a mean $CS_{x} \approx$ 90\% which signifies the percentage of ground truth that lies within $2\sigma$ confidence interval of the predictions (Figure \ref{Confidence Interval}a).  However, for long-term forecasts beyond 4.8 seconds, the CNN-LSTM model (light maroon) outperforms the other two models with a mean  $CS_{x} \approx$ 80\%. Similar trend can be observed for $CS_{y}$ for both short-term and long-term forecasts (Figure \ref{Confidence Interval}b). However, the  confidence score along y is less than x for each model. This shows the model is more confident and captures uncertainty effectively along the direction of predominant motion and suitable for long-term linear motions. However, when the motion is  significantly small along any direction, the model is less confident and uncertainty is not captured accurately.  Among all models, the 1D-CNN probabilistic model (blue) has the lowest confidence score along both x and y across all time horizons. Our results thus indicate that the LSTM model (gray) performs better for short-term probabilistic predictions while CNN-LSTM model (light maroon) has better long-term probabilistic accuracy. Further, the confidence score along y gradually decreased with prediction horizon for each model. This shows the positional accuracy along y for predicted states becomes more uncertain with increase in prediction horizon.  To understand how prediction horizon affects uncertainty estimates, we studied its effect  on performance metrics like ADE and FDE.   Further, we compared each probabilistic model against its deterministic prediction. We also studied the effect of stochastic dropout probability, p  on  performance metrics.

\subsection{Dropout} \label{dropout}

 Stochastic dropout  during inference  is  critical  for inducing  uncertainty  into the model. The network weights are randomly dropped  with  certain probability, p at  every inference generating a new trajectory. The whole inference process was repeated multiple times generating a distribution of  trajectories  (Figure \ref{LSTM + MC Dropout}). For implementing probabilistic inference, we considered dropout layers in  each neural network architecture. Meanwhile the deterministic predictions are obtained using the same  neural network architecture that uses dropout layers only during training but not for inference. For the experiments, each probabilistic model was studied  with dropout probabilities,  p =$\{0.2,0.3,0.4,0.5\}$.   Our main aim is to quantify uncertainty and understand the effect of  dropout probability, p on performance metrics for each probabilistic model and compare that to deterministic prediction. 
 
 \begin{figure}[h!]
  \centering

  \begin{tabular}{@{}c@{}}
    \includegraphics[width=0.9\linewidth, trim = 0cm 1.65cm 0cm 1.75cm, clip]{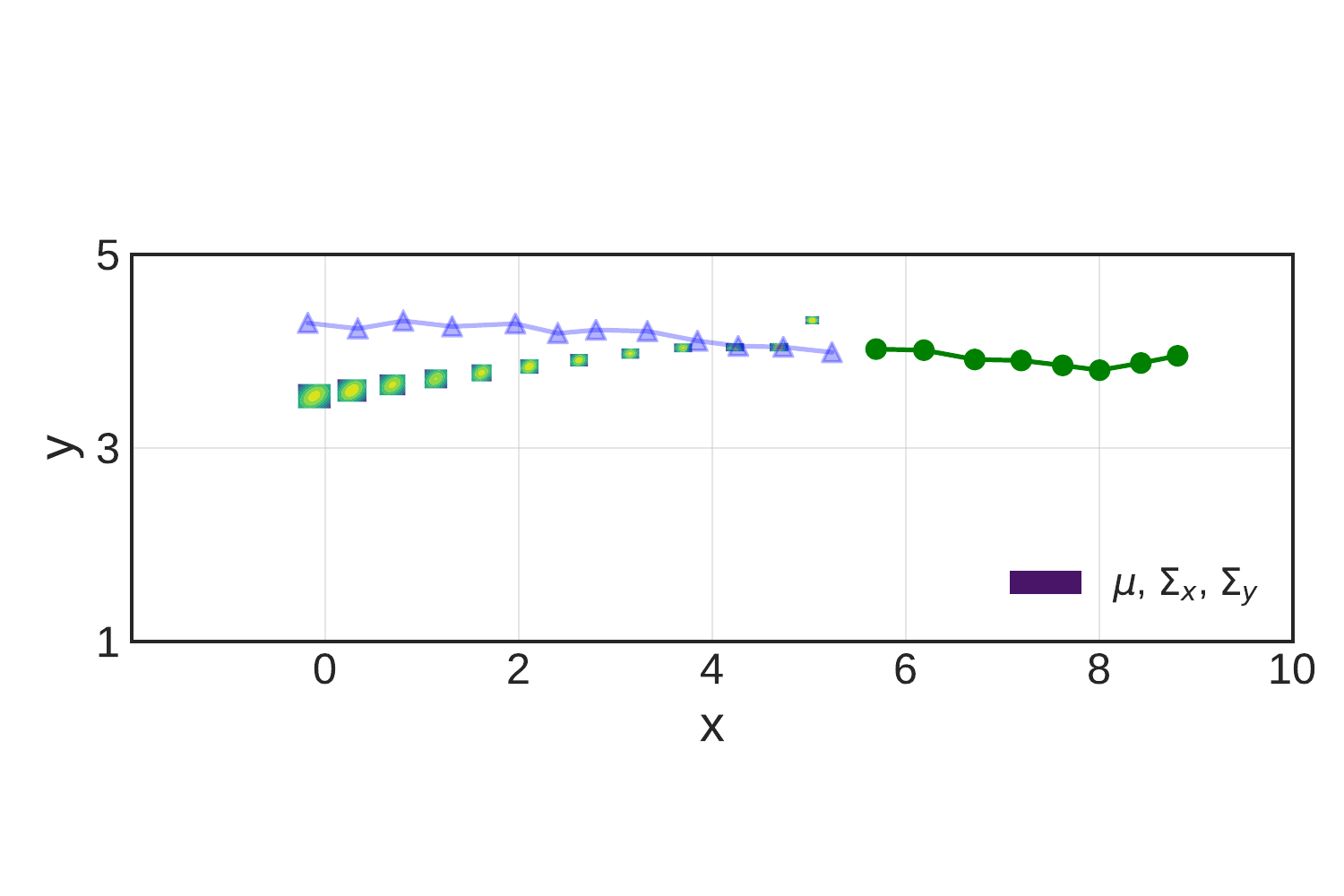} \\[\abovecaptionskip]
    \small (a) 
  \end{tabular}
  
    \begin{tabular}{@{}c@{}}
    \includegraphics[width=0.9\linewidth,  trim = 0cm 1.65cm 0cm 1.75cm, clip]{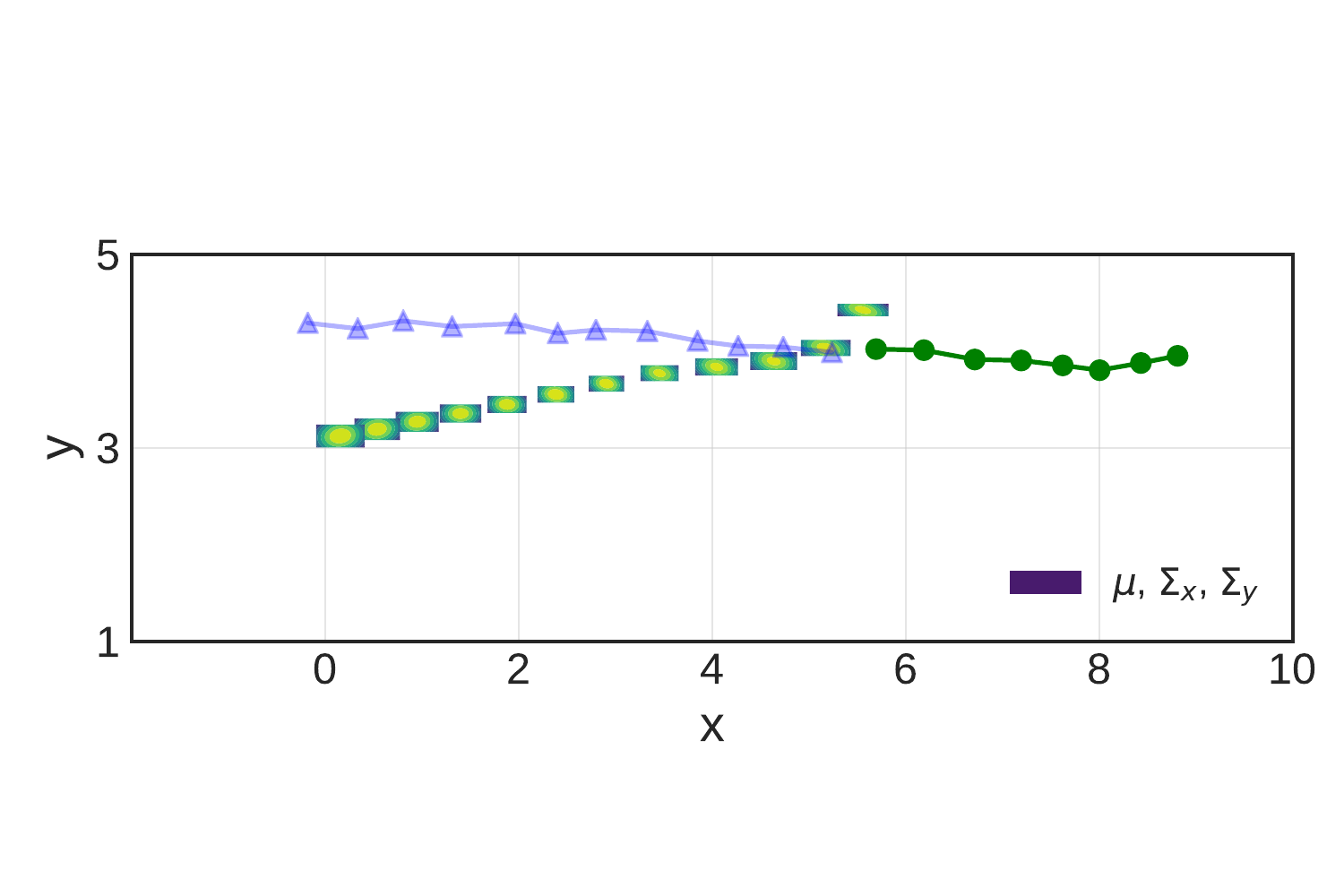} \\[\abovecaptionskip]
    \small (b) 
  \end{tabular}

  \caption{Uncertainty with dropout probability (a) p = 0.2  and (b) p = 0.4  for $T_{f}$ = 4.8 seconds into future  }\label{fig:Dropout_CNN_LSTM_uncertainty}
\end{figure}

In Figure \ref{fig:Dropout_CNN_LSTM_uncertainty}, we compare the uncertainty estimation for dropout probability, p = 0.2 and 0.4 using probabilistic CNN-LSTM model for a particular trajectory.  The results indicate that with increase in dropout probability, the uncertainty in pedestrian state along x increased significantly.  Since, more weights are randomly dropped, the variance associated with prediction also increases and thus the model becomes less certain during prediction. However, no significant difference in uncertainty was observed along y. Further, the mean predicted path is farther away from the ground truth at p=0.4 than at p =0.2 for the current trajectory. A more detailed analysis on variation of performance metrics with stochastic dropout has been provided below.   

\begin{figure}[h!]
  \centering
  \begin{tabular}{@{}c@{}}
    \includegraphics[width=.8\linewidth]{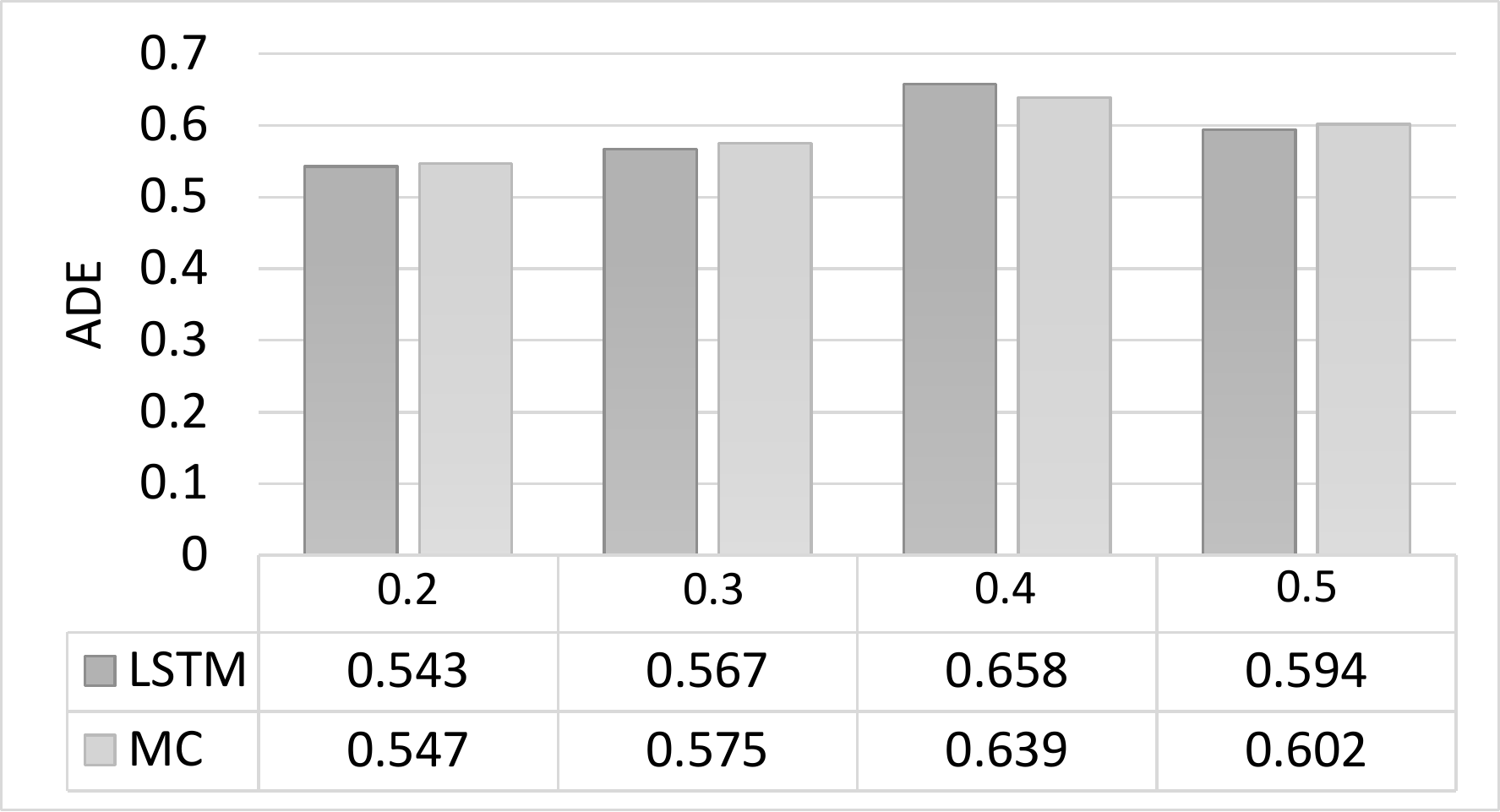} \\[\abovecaptionskip]
  \end{tabular}

  \begin{tabular}{@{}c@{}}
    \includegraphics[width=.8\linewidth]{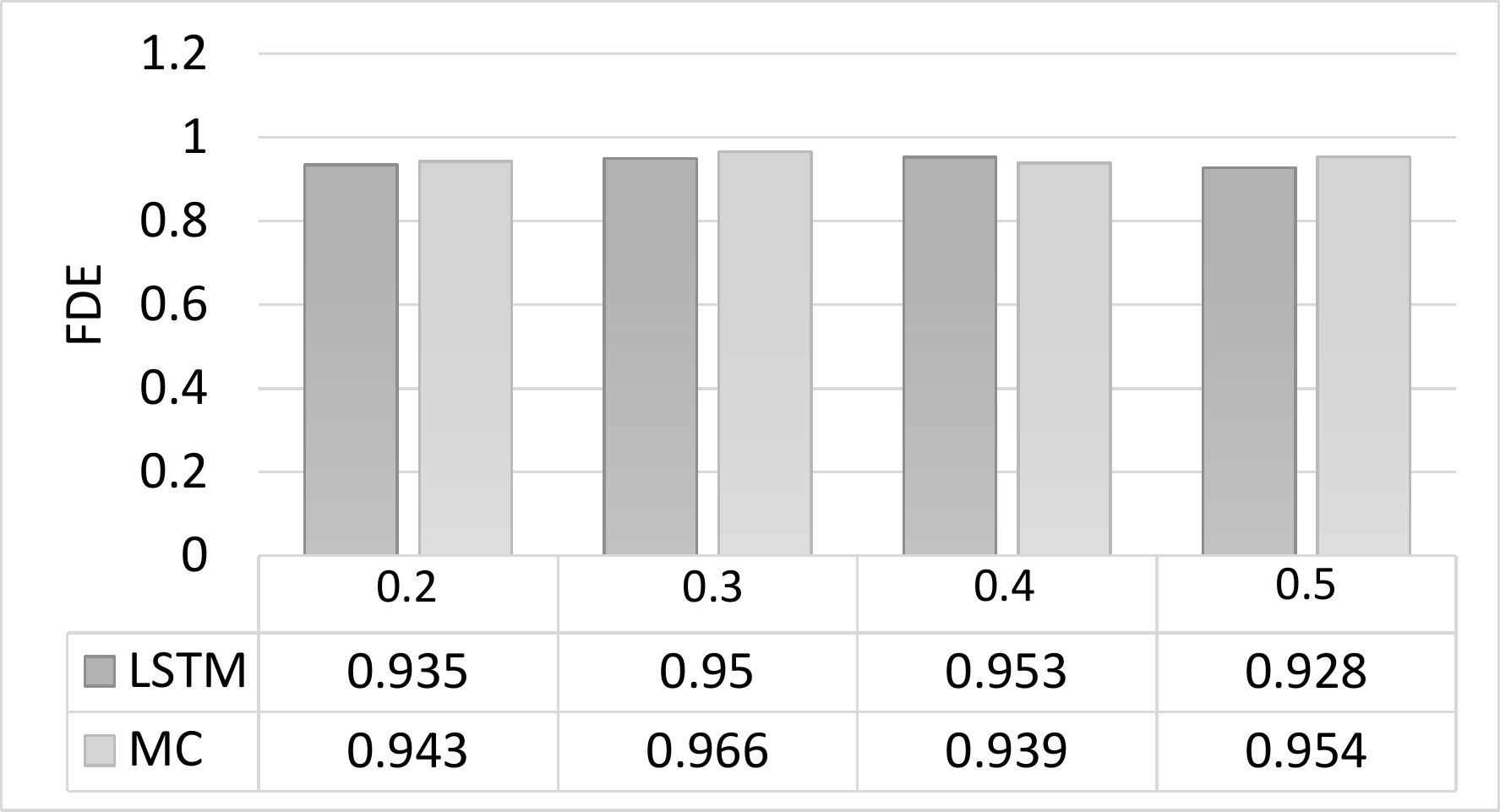} \\[\abovecaptionskip]
  \end{tabular}

  \caption{Performance comparison between  LSTM and LSTM with MC dropout with  varying stochastic dropout. }\label{fig:dropout_LSTM}
\end{figure}

 In Figure \ref{fig:dropout_LSTM},  minimum average displacement error, ADE=0.543 (less is better) was obtained at p=0.2.   Further, the ADE increased with  dropout probability till p=0.4.  It is evident as a higher dropout implies more weights are randomly dropped from the architecture so the variance in trajectories would increase. This increase in randomness across the predicted distribution somehow generates a mean predicted path that has more ADE with the ground truth.   Yet, further increase in dropout probability, p = 0.5 resulted in smaller ADE. Meanwhile, no significant variation was observed in final displacement error with change in dropout probability  for the LSTM model.
 
\begin{figure}[h!]
  \centering
    \begin{tabular}{@{}c@{}}
    \includegraphics[width=.8\linewidth]{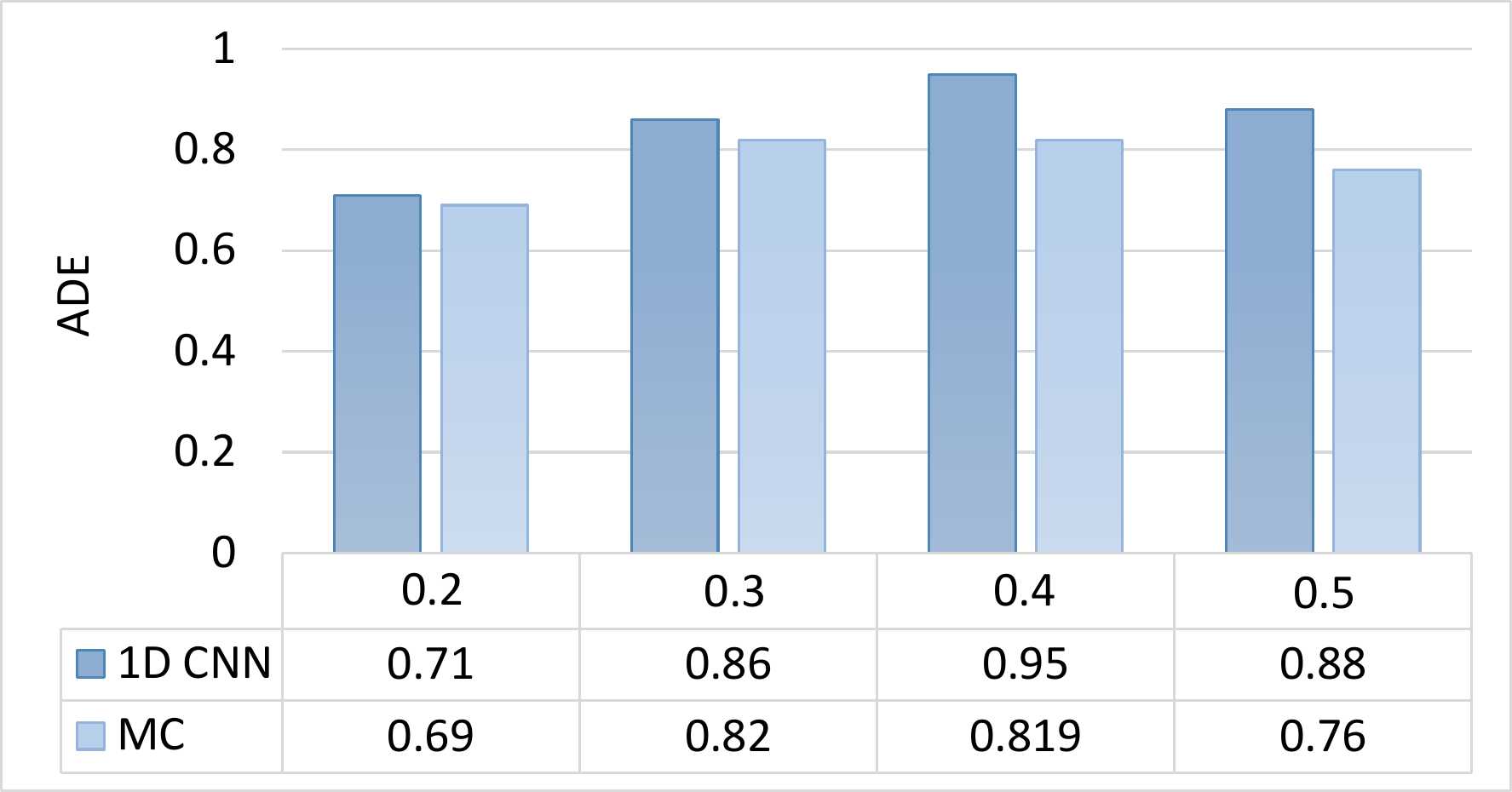} \\[\abovecaptionskip]
  \end{tabular}

  \begin{tabular}{@{}c@{}}
    \includegraphics[width=.8\linewidth]{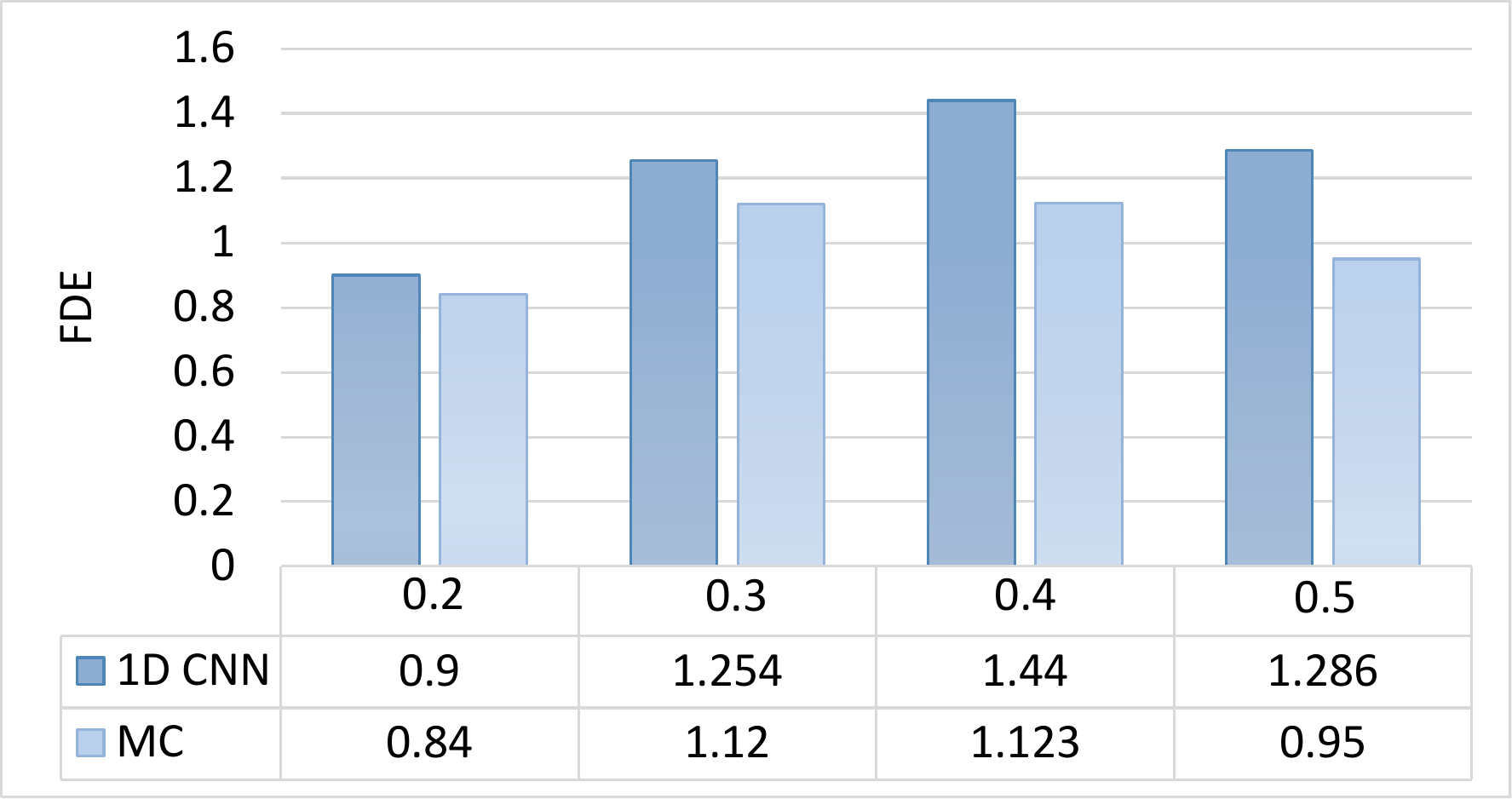} \\[\abovecaptionskip]
  \end{tabular}

  \caption{Performance comparison between  1D CNN and 1D CNN with MC dropout with  varying stochastic dropout. }\label{fig:dropout_1D_CNN}
\end{figure}

\begin{figure}[h!]
  \centering
    \begin{tabular}{@{}c@{}}
    \includegraphics[width=.8\linewidth]{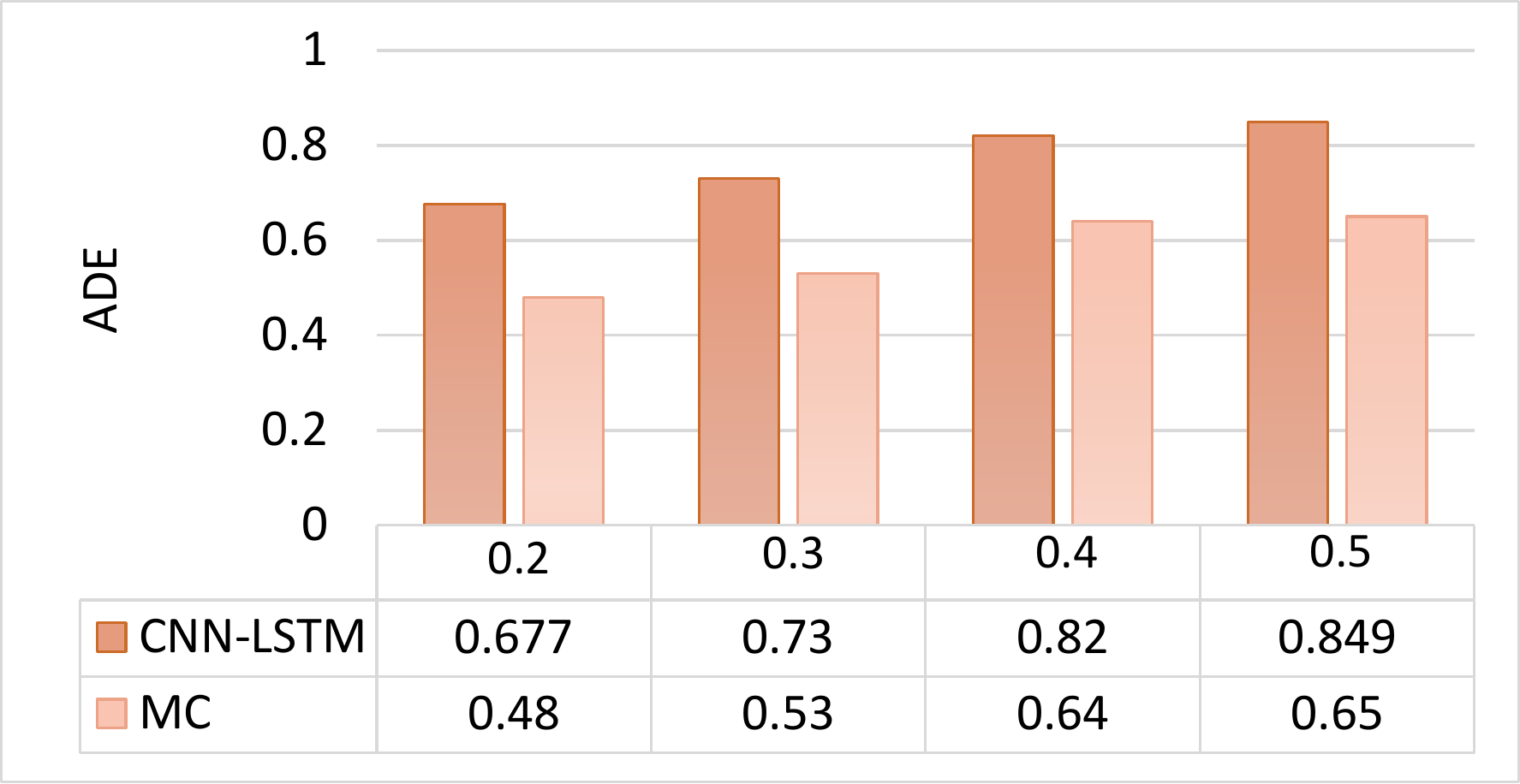} \\[\abovecaptionskip]
  \end{tabular}

  \begin{tabular}{@{}c@{}}
    \includegraphics[width=.8\linewidth]{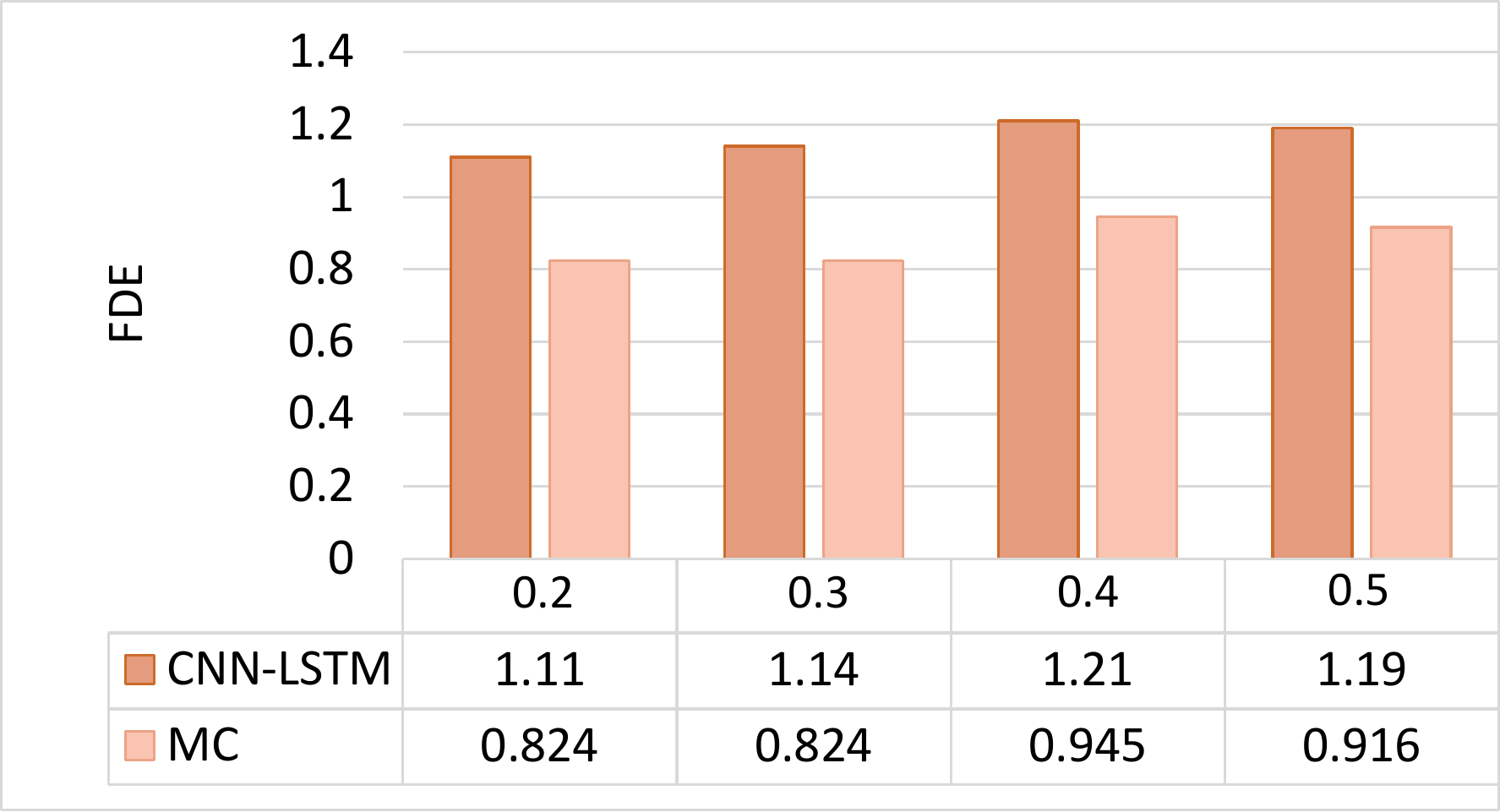} \\[\abovecaptionskip]
  \end{tabular}

  \caption{Performance comparison between  CNN-LSTM and  CNN-LSTM with MC dropout with  varying stochastic dropout. }\label{fig:dropout_CNN_LSTM}
\end{figure}

However, probabilistic predictions with MC dropout on 1D CNN and CNN-LSTM showed significant improvement in ADE and FDE over deterministic predictions across all dropout values.  For instance, at p = 0.4, 1D CNN model with LSTM dropout has an ADE = 0.819 which shows an improvement of \textbf{13.8\%} over vanilla 1D-CNN model with ADE = 0.95 (Figure  \ref{fig:dropout_1D_CNN}). The performance improvement was even higher for the probabilistic CNN-LSTM model with at least \textbf{20\%} improvement in ADE over vanilla CNN-LSTM model (Figure \ref{fig:dropout_CNN_LSTM}) across all dropout values. Similar trend was also seen for the final displacement error where the FDE increased with dropout probability till p=0.4 and then decreased across all models.

Overall,  the probabilistic models 1D-CNN and CNN-LSTM showed significant improvement in both ADE and FDE over deterministic predictions across all dropout probability values. Further, the ADE and FDE increased till p =0.4 across all models which shows dropout induces uncertainty into the model as higher dropout rates  should lead to more variance in trajectory distribution.   However,  further increase in dropout, p = $\{0.5\}$ resulted in  smaller ADE and FDE  (less is better) which seems counter-intuitive.

\subsection{Time Horizon}
We compare the probabilistic and deterministic  models  for multiple time horizons into future.  Based on a past trajectory of T = 3.2 seconds, we predicted the future states along with their associated uncertainty at $T_{f}$ = ${3.2,4.8,6.4,8}$ seconds.  A constant dropout probability, p = 0.2 was considered for all experiments as it had the minimum average displacement error and final displacement error (\ref{dropout}).   Our results indicate that minimum ADE and FDE occur for the smallest time horizon at $T_{f}$ = 3.2 seconds.  Further, both ADE and FDE increased with prediction horizon across all models. It shows that irrespective of the models, error in prediction increases with the increase in prediction horizon. 

\begin{figure}[h!]
  \centering

  \begin{tabular}{@{}c@{}}
    \includegraphics[width=0.9\linewidth, trim = 0cm 1.65cm 0cm 1.75cm, clip]{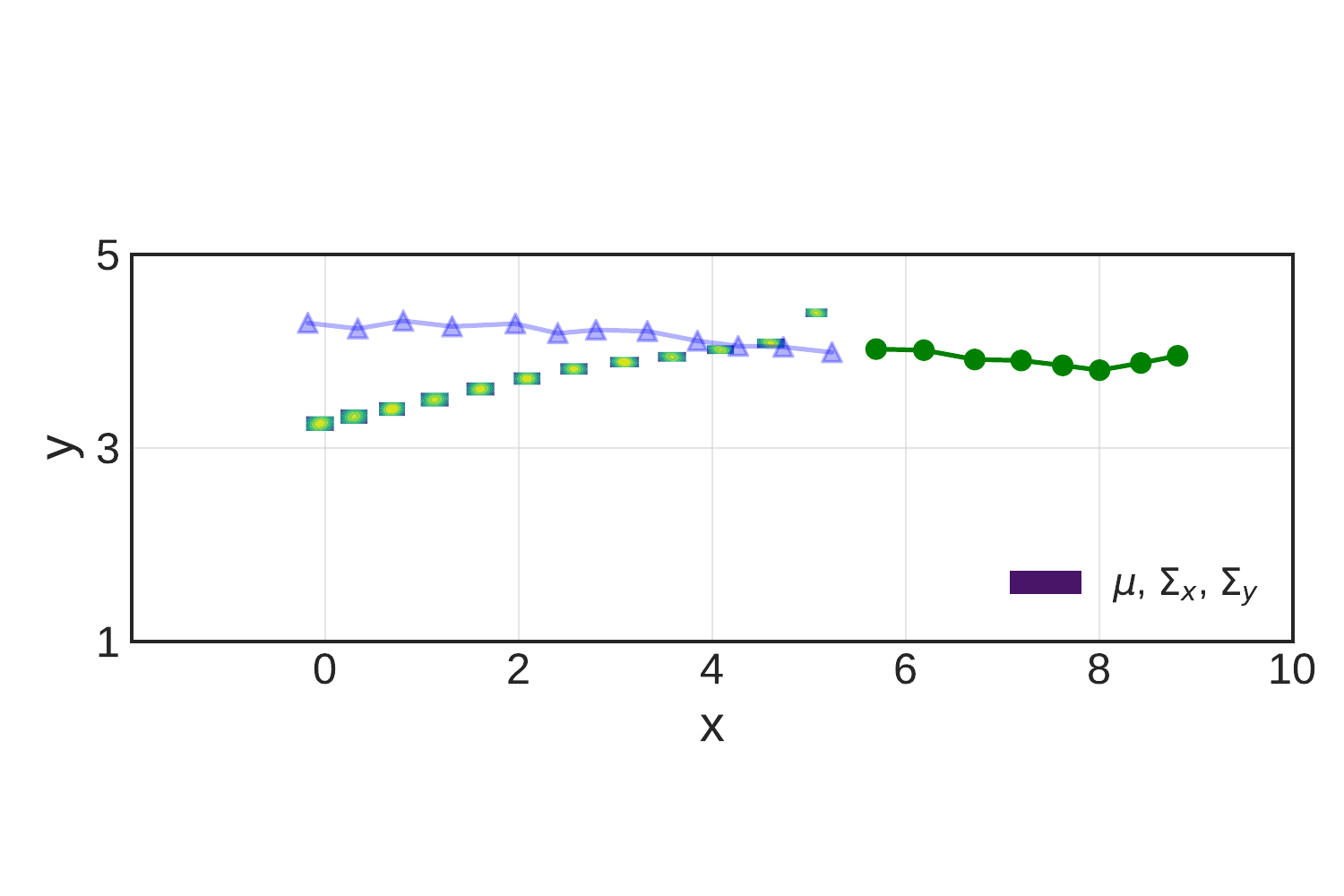} \\[\abovecaptionskip]
     \small (a) 
  \end{tabular}
  
    \begin{tabular}{@{}c@{}}
    \includegraphics[width=0.9\linewidth,  trim = 0cm 1.65cm 0cm 1.75cm, clip]{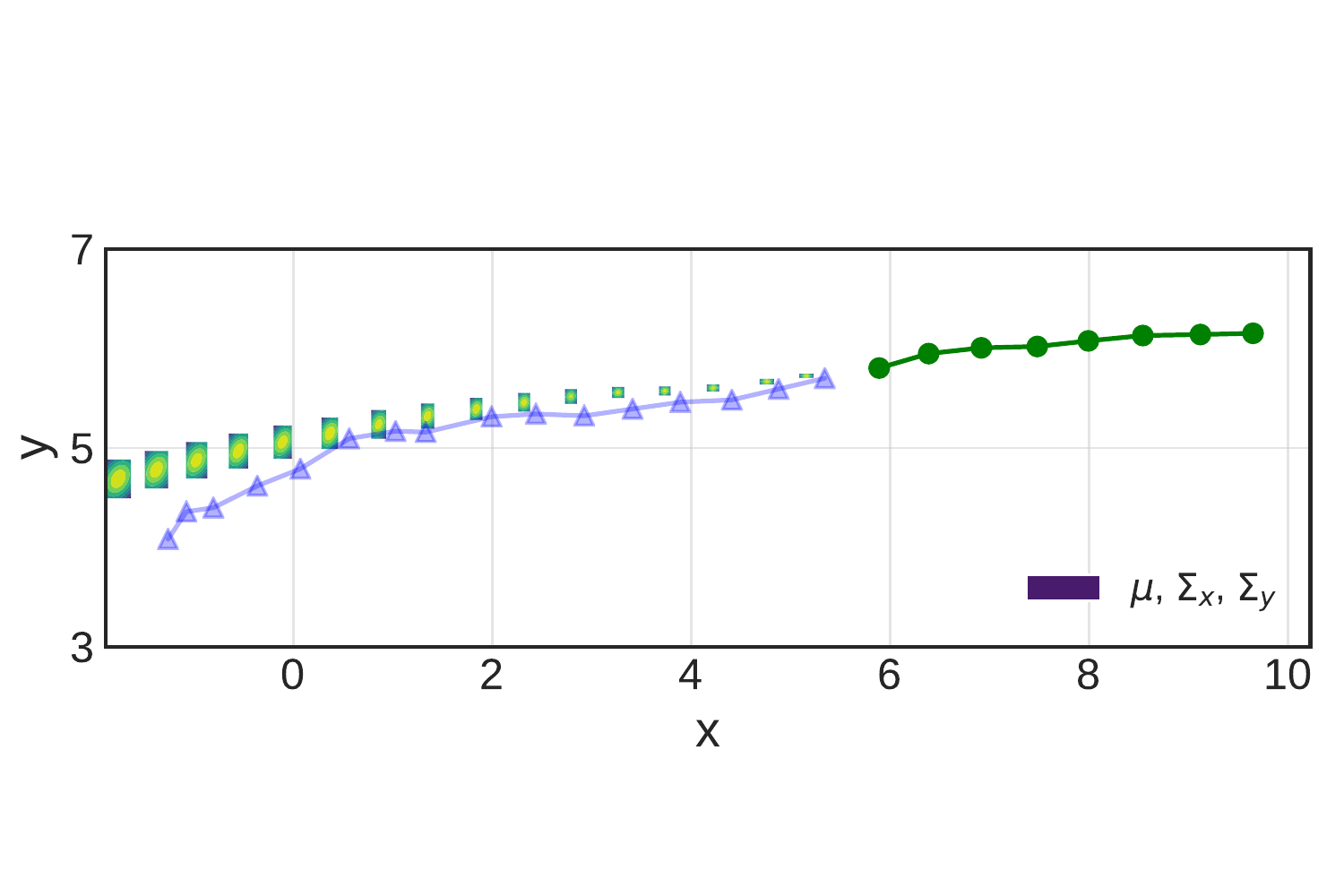} \\[\abovecaptionskip]
      \small (b) 
  \end{tabular}

  \caption{Uncertainty at (a) T = 4.8 secs and (b) T = 6.4 secs with  varying prediction horizon }\label{fig:Time_CNN_LSTM_uncertainty}
\end{figure}

In Figure \ref{fig:Time_CNN_LSTM_uncertainty}, we compare  the uncertainty 
in predicted future states at $T_{f}$ = 3.2 and 4.8 seconds respectively.  Both plots indicate that the uncertainty grows with prediction horizon. Especially, at $T_{f}$ = 4.8 seconds, the variance along both x and y grow significantly with each predicted step when compared with the predicted trajectory at $T_{f}$ = 3.2 seconds. Further, we have  compared the performance metrics based on the mean of predicted trajectories from probabilistic prediction with its deterministic forecast. For both probabilistic and deterministic models, both ADE and FDE increased with prediction horizon.

\begin{figure}[h!]
  \centering
  \begin{tabular}{@{}c@{}}
    \includegraphics[width=.8\linewidth]{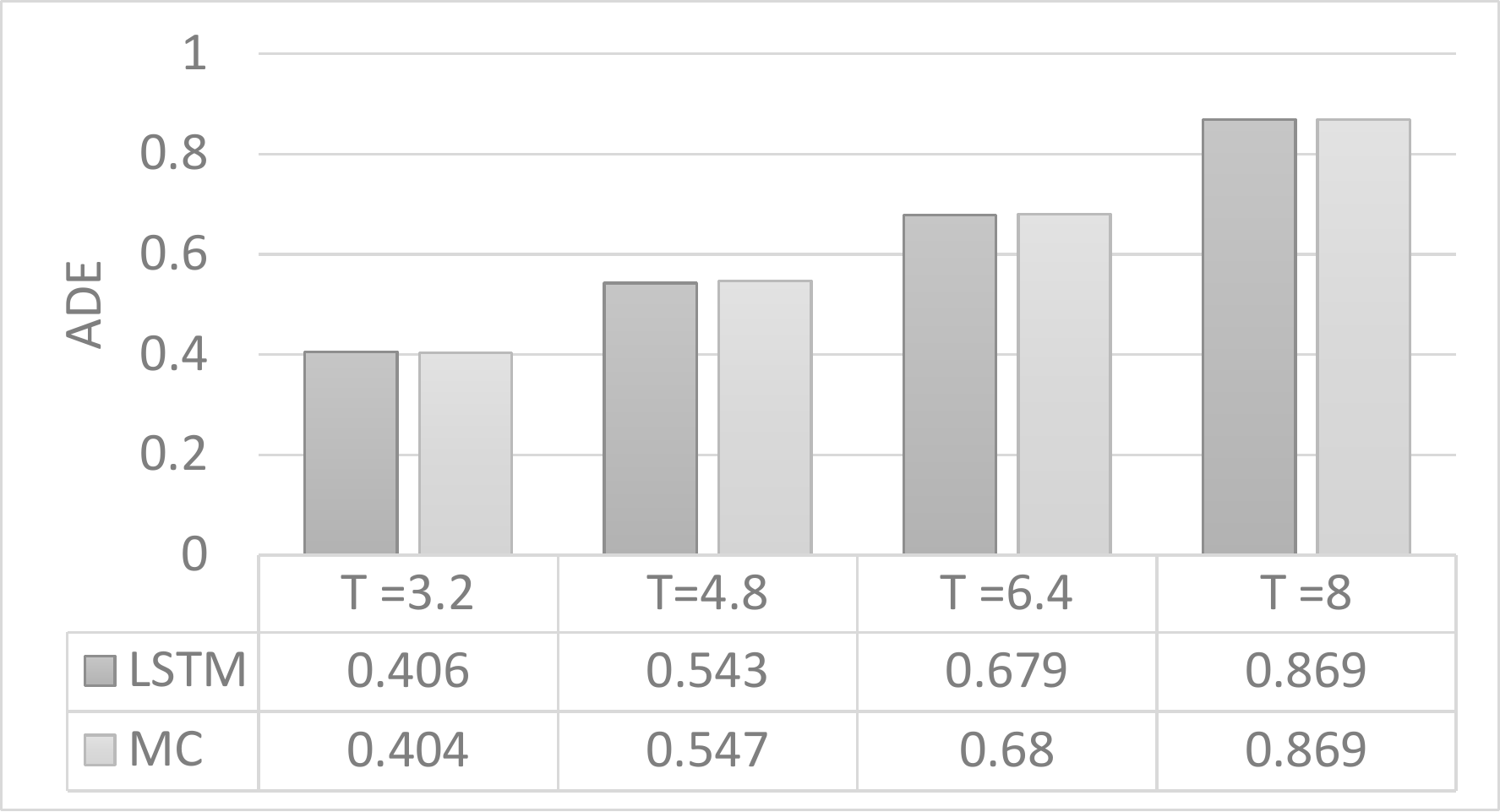} \\[\abovecaptionskip]
  \end{tabular}

  \begin{tabular}{@{}c@{}}
    \includegraphics[width=.8\linewidth]{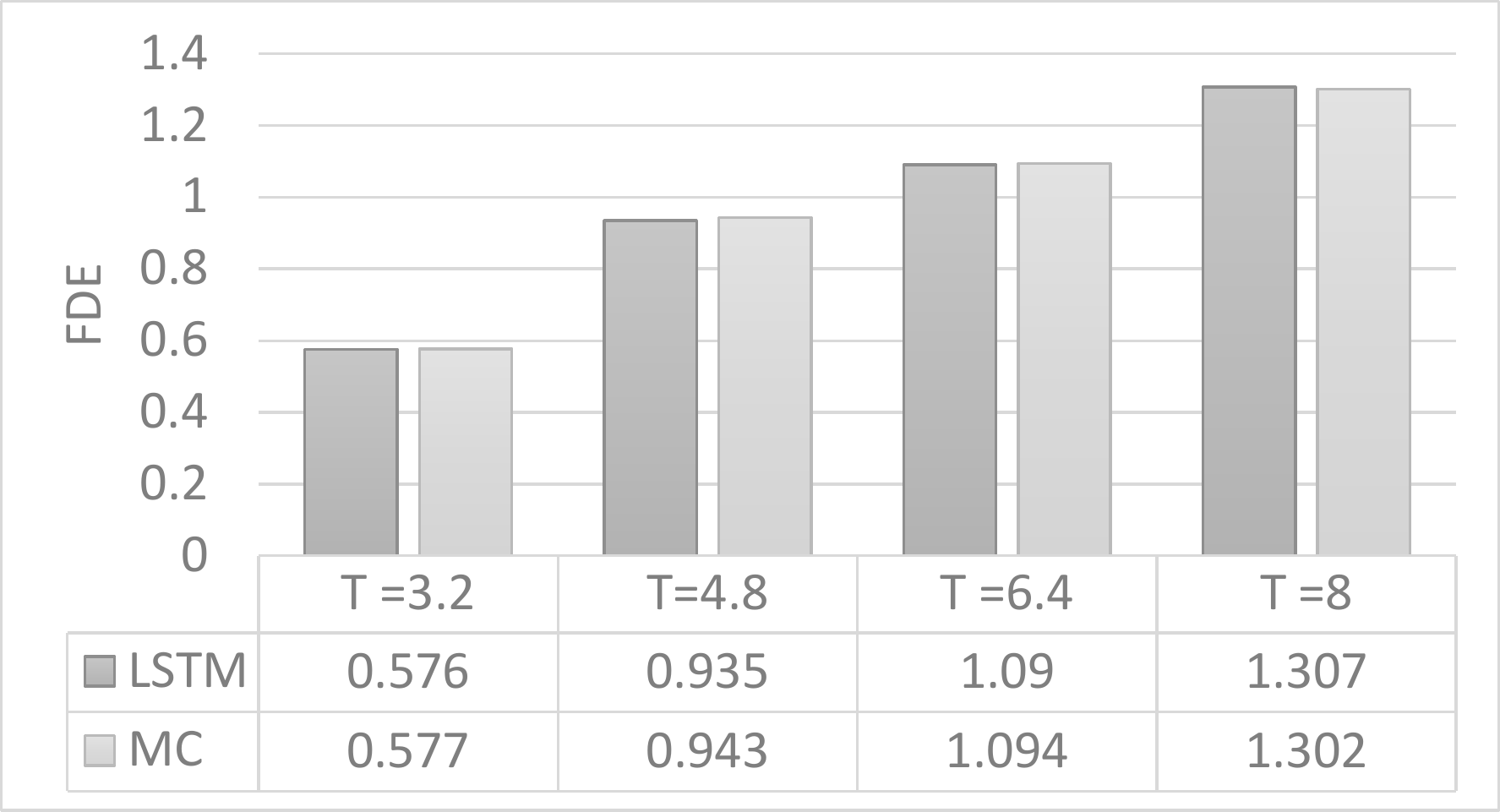} \\[\abovecaptionskip]
  \end{tabular}

  \caption{Performance comparison between  LSTM and  LSTM with MC dropout with  varying prediction horizon }\label{fig:Time_LSTM}
\end{figure}

\begin{figure}[h!]
  \centering
    \begin{tabular}{@{}c@{}}
    \includegraphics[width=.8\linewidth]{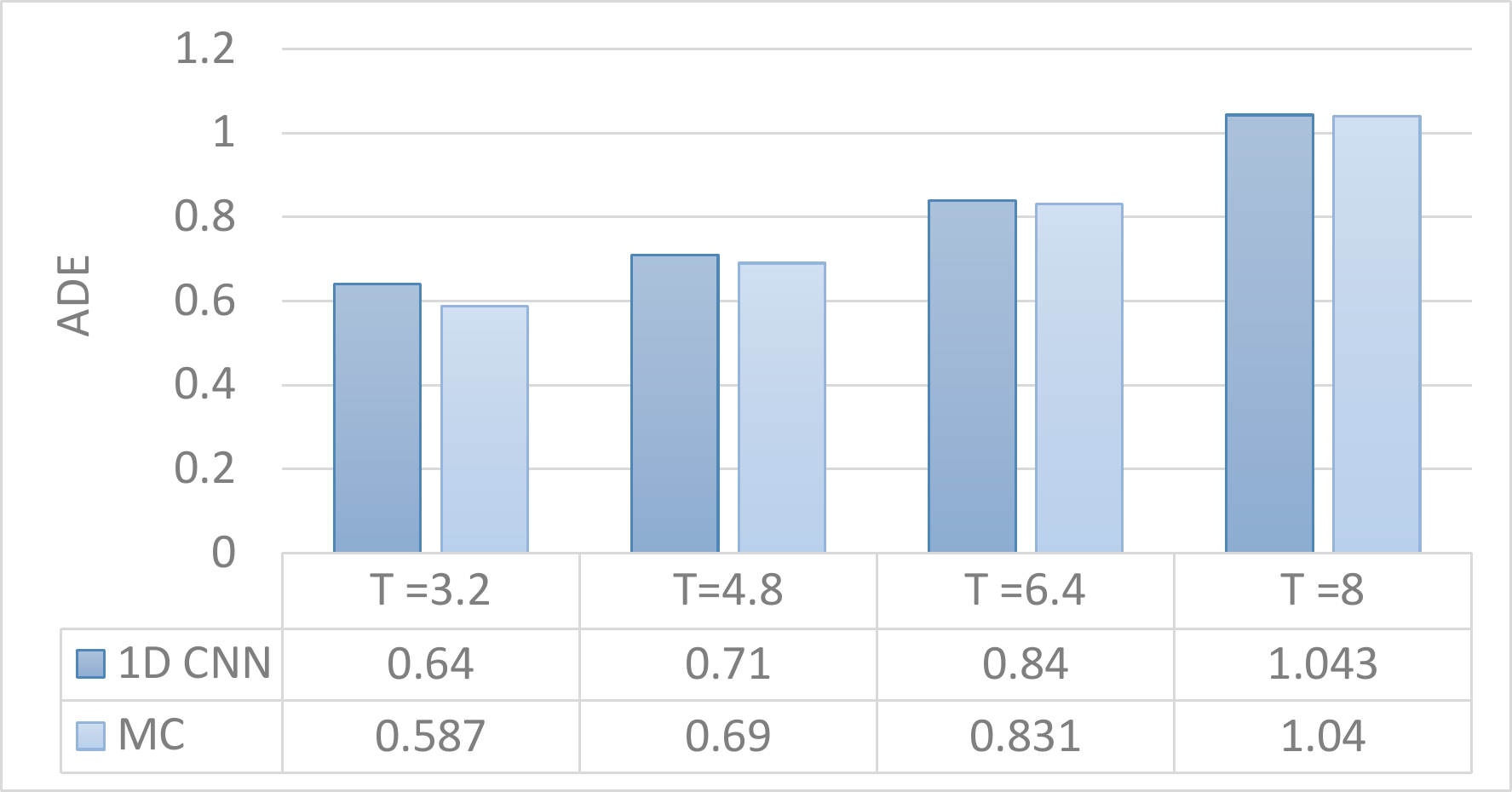} \\[\abovecaptionskip]
  \end{tabular}

  \begin{tabular}{@{}c@{}}
    \includegraphics[width=.8\linewidth]{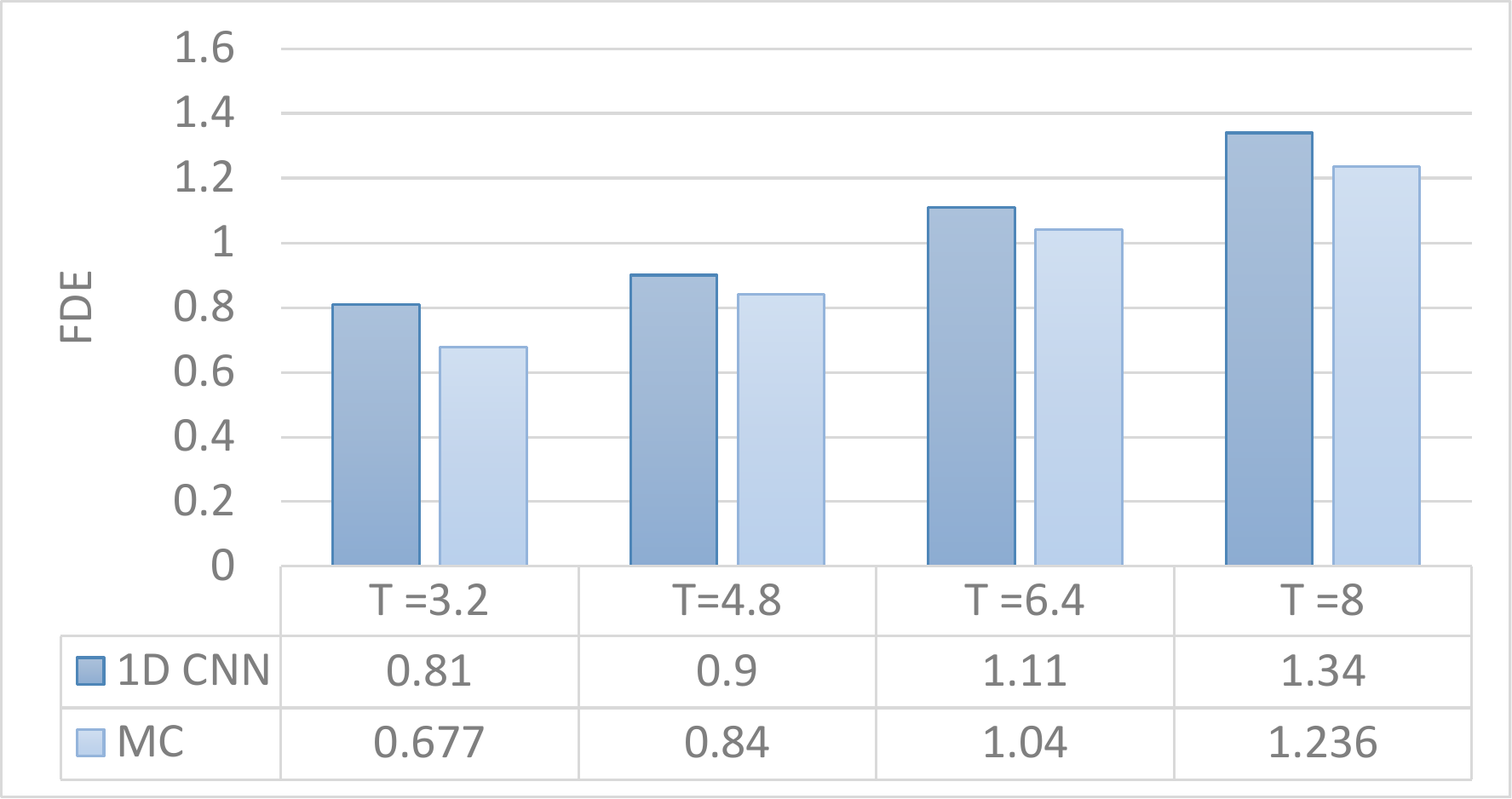} \\[\abovecaptionskip]
  \end{tabular}

  \caption{Performance comparison between  1D CNN and  1D CNN with MC dropout with  varying prediction horizon }\label{fig:Time_1D_CNN}
\end{figure}

\begin{figure}[h!]
  \centering
    \begin{tabular}{@{}c@{}}
    \includegraphics[width=.8\linewidth]{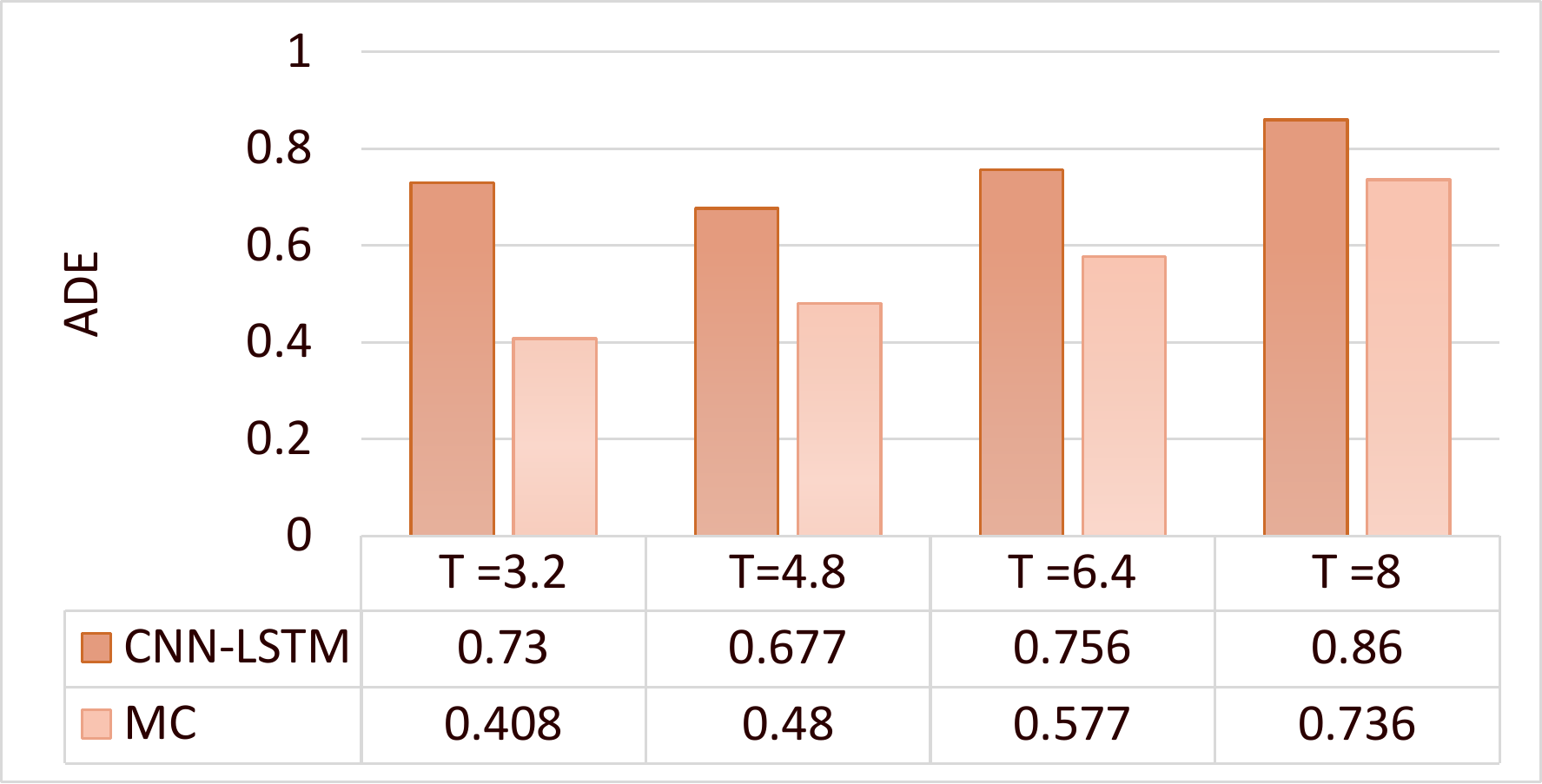} \\[\abovecaptionskip]
  \end{tabular}

  \begin{tabular}{@{}c@{}}
    \includegraphics[width=.8\linewidth]{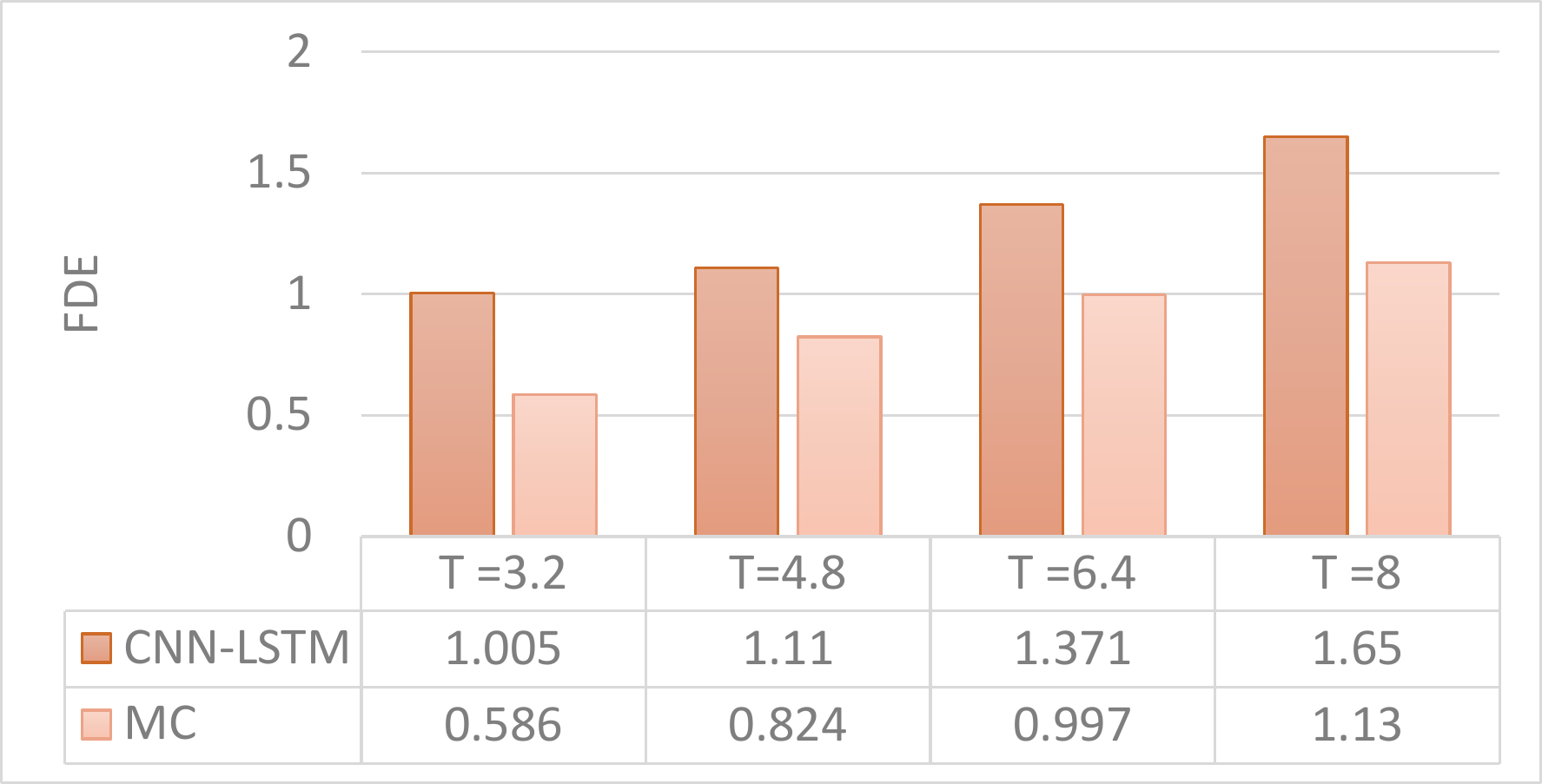} \\[\abovecaptionskip]
  \end{tabular}

  \caption{Performance comparison between CNN-LSTM and CNN-LSTM with MC dropout with  varying prediction horizon }\label{fig:Time_CNN_LSTM}
\end{figure}

  In Figure \ref{fig:Time_LSTM}, LSTM with MC dropout model shows no improvement  in performance metrics over vanilla LSTM. However,  both 1D CNN and CNN-LSTM with MC dropout produce mean predictions that have lower ADE and FDE when compared with the standard deterministic architectures without MC dropout (Figure \ref{fig:Time_1D_CNN},\ref{fig:Time_CNN_LSTM}). Especially, the probabilistic CNN-LSTM model has the lowest ADE and FDE among all probabilistic models across each prediction horizon. Further, the CNN-LSTM model with MC dropout produced atleast \textbf{15\%} and \textbf{35\%} improvement in ADE and FDE respectively over its deterministic prediction.

Overall,  uncertainty increased with both dropout and prediction horizon. Especially, with increase in dropout, the uncertainty increased along the predominant direction of motion i.e x. Meanwhile, increase in prediction horizon lead to increase in uncertainty along lateral motion i.e. y. Further, both dropout and prediction horizon also play an important role on performance metrics.  The mean predicted path from probabilistic models produced a  better estimation of predicted trajectory and is more trust-worthy when compared to a deterministic prediction.

\subsection{Quantitative Evaluation} \label{datasets}

\begin{table}[h!]
\caption{Quantitative ADE/FDE for  predicting 12 future time steps given 8 previous time steps. }
\centering
\begin{tabular}{lccc}
\hline
& ETH        &ZARA1       & ZARA2  \\
\hline
    & \multicolumn{1}{l}{} &\multicolumn{1}{l}{}  & \multicolumn{1}{l}{} \\
LSTM          & 0.54/0.94            & 0.51/0.96            & \textbf{0.53/0.96}   \\
1D CNN        & 0.71/0.90            & 0.75/1.02            & 0.86/1.16            \\
CNN-LSTM      & 0.68/1.11            & 0.73/0.99            & 0.95/1.27            \\
LSTM + MC     & 0.55/0.94            & 0.51/0.96            & 0.54/0.96            \\
1D CNN + MC   & 0.69/0.84            & 0.73/0.99            & 0.85/1.15            \\
CNN-LSTM + MC & \textbf{0.48/0.82}   & \textbf{0.50/0.83}   & 0.77/1.12\\
\hline

\end{tabular}
\label{table:1}
\end{table}

In Table \ref{table:1}, we compare the performance metrics, ADE/FDE based on our predictions for all the models. There are three probabilistic models trained with MC dropout and three standard neural network architectures  which generate deterministic predictions. All the probabilistic models are inferred with a dropout probability, p =0.2. 

We observe that the novel  CNN-LSTM model
with MC dropout outperforms every other model across ETH and ZARA1 dataset. Its mean predicted path  has minimum ADE/FDE (0.48/0.82) in ETH scene. We speculate that the CNN captures feature more efficiently than standard LSTM encoder. Meanwhile, the LSTM decoder utilises the feature information efficiently to generate predictions without any contextual cues (social pooling or scene information). However, the LSTM still performs better in the ZARA2 scene. Unlike other probabilistic models, we did not find any significant effect of MC dropout on predictions for the LSTM model over the standard deterministic model.

\section{CONCLUSIONS AND FUTURE RESEARCH}\label{conclusion}

In this paper, we presented a Bayesian approach using Monte Carlo dropout to quantify the uncertainty in pedestrian trajectory prediction. The method was evaluated on real-world  pedestrian dataset to generate a distribution of trajectories instead of a single trajectory. The current  results indicate that the mean predicted path of probabilistic  model is better and closer to the ground truth than the predictions from deterministic models. Further, the effect of varying dropout probability and time horizon showed that both ADE and FDE increased. It implies the probabilistic models become less certain in its predictions with increase in either prediction horizon or dropout probability. However, the performance metrics of probabilistic models were better than deterministic models. 

In future, we plan to improve the  probabilistic predictions such that the ground truth should always lie within the predicted trajectory distribution. We will explore other Bayesian methods for uncertainty quantification or change the current neural network architecture \cite{zhang_ce2} for accurate long-term prediction.

\addtolength{\textheight}{-2cm}   




\section*{ACKNOWLEDGMENT}

This project was funded in part by the Safety through Disruption (Safe-D) National University Transportation Center (UTC), a grant from the U.S. Department of Transportation – Office of the Assistant Secretary for Research and Technology, University Transportation Centers Program.

	\begin{IEEEbiography}[{\includegraphics[width = 1 in, height = 1.25 in, clip, keepaspectratio]{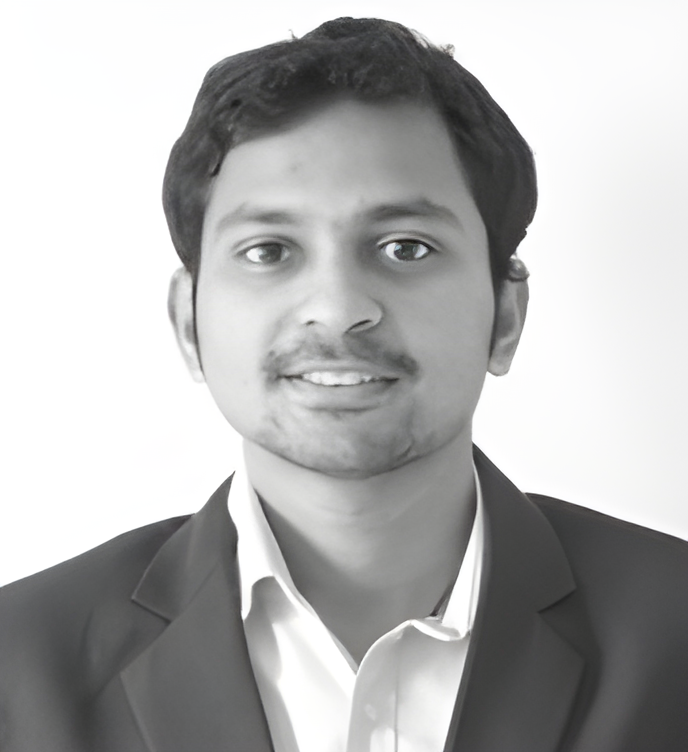}}]{Anshul Nayak}
	
		received his B.Tech  in mechanical engineering from NIT, Rourkela, India. He completed his Master's degree in Mechanical engineering at Virginia Tech and is currently pursuing his Ph.D at the  Autonomous Systems and Intelligent Machines (ASIM) lab at the same university. His research interests include cooperative planning and uncertainty estimation in prediction.
		
	\end{IEEEbiography}


\begin{IEEEbiography}[{\includegraphics[width = 1 in, height = 1.25 in, clip, keepaspectratio]{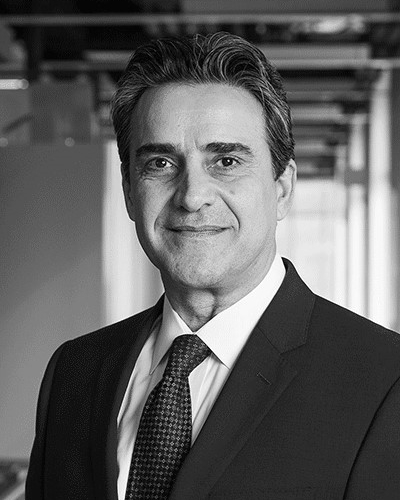}}]{Azim Eskandarian}
	
		has been a Professor and Head of the Mechanical Engineering Department at Virginia Tech since August 2015. He became the Nicholas and Rebecca Des Champs chaired Professor in April 2018. He also has a courtesy appointment as a Professor in the Electrical and Computer Engineering Department since 2021. He established the Autonomous Systems and Intelligent Machines laboratory at Virginia Tech and has conducted pioneering research in autonomous vehicles, human/driver cognition and vehicle interface, advanced driver assistance systems, and robotics. Before joining Virginia Tech, he was a Professor of Engineering and Applied Science at George Washington University (GWU) and the Founding Director of the Center for Intelligent Systems Research, from 1996 to 2015, the Director of the Transportation Safety and Security University Area of Excellence, from 2002 to 2015, and the Co-Founder of the National Crash Analysis Center in 1992 and its Director from 1998 to 2002 and 2013 to 2015. From 1989 to 1992, he was an Assistant Professor at Pennsylvania State University, York, PA, and an Engineer/Project Manager in the industry from 1983 to 1989. Dr. Eskandarian is a Fellow of ASME, a member of SAE, and a Senior Member of IEEE professional societies. He received SAE’s Vincent Bendix Automotive Electronics Engineering Award in 2021, IEEE ITS Society’s Outstanding Researcher Award in 2017, and GWU’s School of Engineering Outstanding Researcher Award in 2013.
			\end{IEEEbiography}

\begin{IEEEbiography}[{\includegraphics[width = 1 in, height = 1.25 in, clip, keepaspectratio]{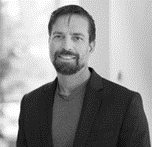}}]{Zachary Doerzaph}
	
	 is the Executive Director of the Virginia Tech Transportation Institute (VTTI), a global leader in transportation research.  Working alongside a talented team, Doerzaph focuses on creating a future of ubiquitous, safe, and effective mobility by conducting innovative and impactful research today.  Also, a faculty member within the Department of Biomedical Engineering and Mechanics at Virginia Tech, Doerzaph works with fellow faculty to provide experiential learning opportunities to prepare the next generation workforce. Doerzaph is known for innovative and extensive transportation research and leadership projects. His work focuses on maximizing performance at the interface of driver, vehicle, and infrastructure systems through the application of advanced technologies.
		
	\end{IEEEbiography}
	
	\vfill
	

\begin{thebibliography}{99}
\bibliographystyle{IEEEtran}

\bibitem{Zhang_ce1}Zhang, Ce, and Azim Eskandarian. "A Quality Index Metric and Method for Online Self-Assessment of Autonomous Vehicles Sensory Perception." arXiv preprint arXiv:2203.02588 (2022).

\bibitem{Nayak}Nayak, Anshul, Azim Eskandarian, Prasenjit Ghorai, and Zachary Doerzaph. "A Comparative Study on Feature Descriptors for Relative Pose Estimation in Connected Vehicles." In ASME International Mechanical Engineering Congress and Exposition, vol. 85628, p. V07BT07A022. American Society of Mechanical Engineers, 2021.

\bibitem{Eskandarian}A. Eskandarian, C. Wu and C. Sun, "Research Advances and Challenges of Autonomous and Connected Ground Vehicles," in IEEE Transactions on Intelligent Transportation Systems, vol. 22, no. 2, pp. 683-711, Feb. 2021, doi: 10.1109/TITS.2019.2958352.

\bibitem{Xue}Xue, Hao, Du Q. Huynh, and Mark Reynolds. "Bi-prediction: pedestrian trajectory prediction based on bidirectional LSTM classification." In 2017 International Conference on Digital Image Computing: Techniques and Applications (DICTA), pp. 1-8. IEEE, 2017.

\bibitem{Hao}Xue, Hao, Du Q. Huynh, and Mark Reynolds. "SS-LSTM: A hierarchical LSTM model for pedestrian trajectory prediction." In 2018 IEEE Winter Conference on Applications of Computer Vision (WACV), pp. 1186-1194. IEEE, 2018.


\bibitem{Houenou}A. Houenou, P. Bonnifait, V. Cherfaoui and W. Yao, "Vehicle trajectory prediction based on motion model and maneuver recognition," 2013 IEEE/RSJ International Conference on Intelligent Robots and Systems, 2013, pp. 4363-4369, doi: 10.1109/IROS.2013.6696982.

\bibitem{Abbeel}Kahn, Gregory, Adam Villaflor, Vitchyr Pong, Pieter Abbeel, and Sergey Levine. "Uncertainty-aware reinforcement learning for collision avoidance." arXiv preprint arXiv:1702.01182 (2017).

\bibitem{Suo}Suo, Yongfeng, Wenke Chen, Christophe Claramunt, and Shenhua Yang. "A ship trajectory prediction framework based on a recurrent neural network." Sensors 20, no. 18 (2020): 5133.

\bibitem{Park}Park, Seong Hyeon, ByeongDo Kim, Chang Mook Kang, Chung Choo Chung, and Jun Won Choi. "Sequence-to-sequence prediction of vehicle trajectory via LSTM encoder-decoder architecture." In 2018 IEEE Intelligent Vehicles Symposium (IV), pp. 1672-1678. IEEE, 2018.

\bibitem{Florent}Altché, Florent, and Arnaud de La Fortelle. "An LSTM network for highway trajectory prediction." In 2017 IEEE 20th international conference on intelligent transportation systems (ITSC), pp. 353-359. IEEE, 2017.

\bibitem{Manh}Manh, Huynh, and Gita Alaghband. "Scene-lstm: A model for human trajectory prediction." arXiv preprint arXiv:1808.04018 (2018).

\bibitem{Alahi}Alahi, Alexandre, Kratarth Goel, Vignesh Ramanathan, Alexandre Robicquet, Li Fei-Fei, and Silvio Savarese. "Social lstm: Human trajectory prediction in crowded spaces." In Proceedings of the IEEE conference on computer vision and pattern recognition, pp. 961-971. 2016.

\bibitem{Nikhil}Nikhil, Nishant, and Brendan Tran Morris. "Convolutional neural network for trajectory prediction." In Proceedings of the European Conference on Computer Vision (ECCV) Workshops, pp. 0-0. 2018.

\bibitem{Simone}Zamboni, Simone, Zekarias Tilahun Kefato, Sarunas Girdzijauskas, Christoffer Norén, and Laura Dal Col. "Pedestrian trajectory prediction with convolutional neural networks." Pattern Recognition 121 (2022): 108252.


\bibitem{Xihui}Wu, Xihui, Anshul Nayak, and Azim Eskandarian. "Motion Planning of Autonomous Vehicles under Dynamic Traffic Environment in Intersections Using Probabilistic Rapidly Exploring Random Tree." SAE International Journal of Connected and Automated Vehicles 4, no. 12-04-04-0029 (2021).


\bibitem{Ammoun}Ammoun, Samer, and Fawzi Nashashibi. "Real time trajectory prediction for collision risk estimation between vehicles." In 2009 IEEE 5th International Conference on Intelligent Computer Communication and Processing, pp. 417-422. IEEE, 2009.

\bibitem{Ellis}Ellis, David, Eric Sommerlade, and Ian Reid. "Modelling pedestrian trajectory patterns with gaussian processes." In 2009 IEEE 12th International Conference on Computer Vision Workshops, ICCV Workshops, pp. 1229-1234. IEEE, 2009.

\bibitem{Wiest}Wiest, Jürgen, Matthias Höffken, Ulrich Kreßel, and Klaus Dietmayer. "Probabilistic trajectory prediction with Gaussian mixture models." In 2012 IEEE Intelligent Vehicles Symposium, pp. 141-146. IEEE, 2012.

\bibitem{Sun}Sun, C., and Eskandarian, A. "A Predictive Frontal and Oblique Collision Mitigation System for Autonomous Vehicles." ASME. Letters Dyn. Sys. Control. October 2021; 1(4): 041012.

\bibitem{Khattar}Khattar, Vanshaj, and Azim Eskandarian. "Stochastic Predictive Control for Crash Avoidance in Autonomous Vehicles Based on Stochastic Reachable Set Threat Assessment." In ASME International Mechanical Engineering Congress and Exposition, vol. 85628, p. V07BT07A026. American Society of Mechanical Engineers, 2021.

\bibitem{Althoff}Althoff, Matthias. "Reachability analysis and its application to the safety assessment of autonomous cars." PhD diss., Technische Universität München, 2010.

\bibitem{Zernetsch}Zernetsch, Stefan, Hannes Reichert, Viktor Kress, Konrad Doll, and Bernhard Sick. "Trajectory forecasts with uncertainties of vulnerable road users by means of neural networks." In 2019 IEEE Intelligent Vehicles Symposium (IV), pp. 810-815. IEEE, 2019.

\bibitem{Kim}Kim, ByeoungDo, Chang Mook Kang, Jaekyum Kim, Seung Hi Lee, Chung Choo Chung, and Jun Won Choi. "Probabilistic vehicle trajectory prediction over occupancy grid map via recurrent neural network." In 2017 IEEE 20th International Conference on Intelligent Transportation Systems (ITSC), pp. 399-404. IEEE, 2017.

\bibitem{Gal} Gal, Yarin, and Zoubin Ghahramani. "Dropout as a bayesian approximation: Representing model uncertainty in deep learning." In international conference on machine learning, pp. 1050-1059. PMLR, 2016.


\bibitem{Zhang}Zhang, Xiaoge, and Sankaran Mahadevan. "Bayesian neural networks for flight trajectory prediction and safety assessment." Decision Support Systems 131 (2020): 113246.

\bibitem{Zhu}Zhu, Lingxue, and Nikolay Laptev. "Deep and confident prediction for time series at uber." In 2017 IEEE International Conference on Data Mining Workshops (ICDMW), pp. 103-110. IEEE, 2017.

\bibitem{Jospin}Jospin, Laurent Valentin, Wray Buntine, Farid Boussaid, Hamid Laga, and Mohammed Bennamoun. "Hands-on Bayesian neural networks--a tutorial for deep learning users." arXiv preprint arXiv:2007.06823 (2020).



\bibitem{Pelligrini}Pellegrini, Stefano, Andreas Ess, and Luc Van Gool. "Improving data association by joint modeling of pedestrian trajectories and groupings." In European conference on computer vision, pp. 452-465. Springer, Berlin, Heidelberg, 2010.
\bibitem{UCY}L. Leal-Taixé, M. Fenzi, A. Kuznetsova, B. Rosenhahn and S. Savarese, "Learning an Image-Based Motion Context for Multiple People Tracking," 2014 IEEE Conference on Computer Vision and Pattern Recognition, 2014, pp. 3542-3549, doi: 10.1109/CVPR.2014.453.
\bibitem{Takens}Takens, Floris. "Detecting strange attractors in turbulence." In Dynamical systems and turbulence, Warwick 1980, pp. 366-381. Springer, Berlin, Heidelberg, 1981.

\bibitem{zhang_ce2}Zhang, Ce, Azim Eskandarian, and Xuelai Du. "Attention-based Neural Network for Driving Environment Complexity Perception." In 2021 IEEE International Intelligent Transportation Systems Conference (ITSC), pp. 2781-2787. IEEE, 2021.

\end{thebibliography}
\end{document}